\documentclass[10pt,twocolumn,letterpaper]{article}

\usepackage[pagenumbers]{iccv} 

\usepackage{amsmath}
\usepackage{graphicx}  
\usepackage{booktabs}   
\usepackage{array}   

\usepackage{multirow}
\usepackage{hhline}
\usepackage{makecell}
\usepackage{bm}
\usepackage{wrapfig}
\usepackage{bbding}
\usepackage{caption}
\usepackage{mathrsfs}
\usepackage{framed}
\usepackage{makecell}
\usepackage{color}
\usepackage{colortbl}  
\usepackage{xcolor}
\usepackage{threeparttable}

\definecolor{iccvblue}{rgb}{0.21,0.49,0.74}
\usepackage[pagebackref,breaklinks,colorlinks,allcolors=iccvblue]{hyperref}

\usepackage[accsupp]{axessibility} 
\usepackage{xcolor}
\definecolor{lightyellow}{RGB}{255, 255, 204}
\definecolor{lightred}{RGB}{255, 204, 204}

\def\paperID{11625} 
\def\confName{ICCV}
\def\confYear{2025}

\title{Fewer Denoising Steps or Cheaper Per-Step Inference: Towards Compute-Optimal Diffusion Model Deployment}

\author{Zhenbang Du\thanks{Equal contribution.} , Yonggan Fu$^{*}$, Lifu Wang$^{*}$, Jiayi Qian, Xiao Luo,\\
Yingyan (Celine) Lin\\
Georgia Institute of Technology\\
    {\tt\small \{zdu89,jiayiqian,celine.lin\}@gatech.edu
    }  
}

\begin{document}
\maketitle
\begin{abstract}

Diffusion models have shown remarkable success across generative tasks, yet their high computational demands challenge deployment on resource-limited platforms. This paper investigates a critical question for compute-optimal diffusion model deployment: Under a post-training setting without fine-tuning, is it more effective to reduce the number of denoising steps or to use a cheaper per-step inference? Intuitively, reducing the number of denoising steps increases the variability of the distributions across steps, making the model more sensitive to compression. In contrast, keeping more denoising steps makes the differences smaller, preserving redundancy, and making post-training compression more feasible.  
To systematically examine this, we propose PostDiff, a \textit{training-free} framework for accelerating pre-trained diffusion models by reducing redundancy at 
both the input level and module level in a post-training manner. At the input level, we propose a mixed-resolution denoising scheme based on the insight that reducing generation resolution in early denoising steps can enhance low-frequency components and improve final generation fidelity. At the module level, we employ a hybrid module caching strategy to reuse computations across denoising steps. Extensive experiments and ablation studies demonstrate that (1) PostDiff can significantly improve the fidelity-efficiency trade-off of state-of-the-art diffusion models, and (2) to boost efficiency while maintaining decent generation fidelity, reducing per-step inference cost is often more effective than reducing the number of denoising steps. Our code is available at \href{https://github.com/GATECH-EIC/PostDiff}{https://github.com/GATECH-EIC/PostDiff}.

\end{abstract}    
\section{Introduction}
\label{sec:intro}

Diffusion models have rapidly advanced as a powerful tool for various generative tasks, such as image synthesis and video generation. Their success stems from the iterative denoising process, which gradually refines outputs to achieve high-quality results, along with advanced model architectures. However, both the iterative nature and model complexity of diffusion models make them computationally intensive, posing a significant challenge for deployment on resource-constrained platforms, where computational resources, memory, and power are limited. Efficient deployment of diffusion models on these platforms is therefore highly desirable to broaden the accessibility of generative AI, enabling real-time applications that can operate within the strict constraints of on-device resources.

With the increasing demand for rapid diffusion model deployment, where a given diffusion model is expected to be instantly compressed and deployed without fine-tuning, a critical question arises: Under a post-training setting, is it more effective to optimize for fewer denoising steps or to reduce the per-step inference cost? This question is particularly important because each approach entails different trade-offs: Reducing the number of denoising steps introduces greater variability in feature changes between steps, which can amplify the negative impact of compression on output fidelity; Conversely, using more denoising steps reduces these step-to-step differences, preserving redundancy and making it easier to apply post-training compression without sacrificing quality. However, there is a limited exploration of how these two strategies perform in terms of fidelity-efficiency trade-offs, and investigating this could provide valuable guidelines for efficient diffusion model deployment in real-world applications.

\begin{figure}
    \centering
    \vspace{-0.5em}
    \includegraphics[width=0.99\linewidth]{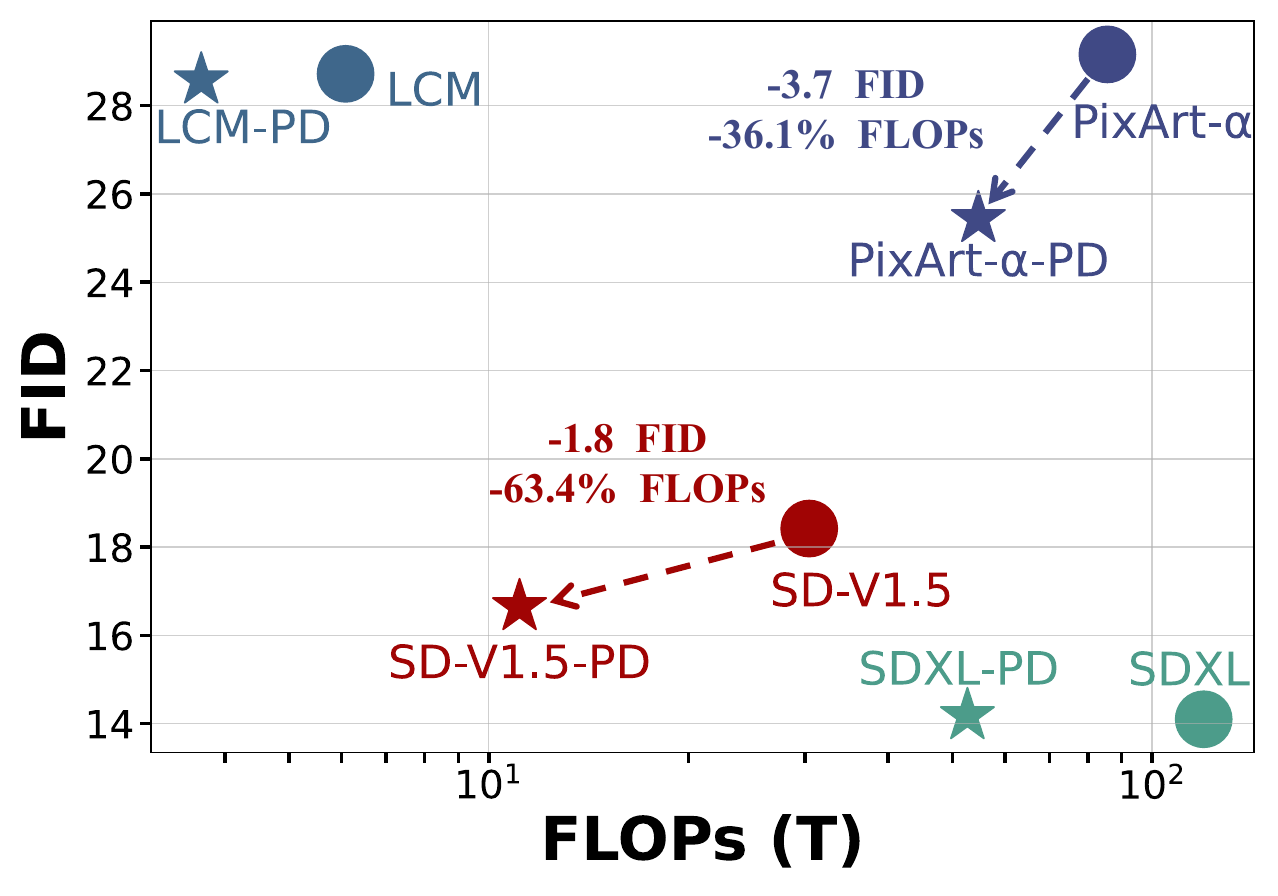}
    \vspace{-1em}
        \caption{Apply our PostDiff technique to SOTA diffusion models, where ``-PD" denotes the models delivered by PostDiff.}
    \label{fig:sota}
    \vspace{-1.5em}
\end{figure}

To address this research question, we propose PostDiff, a \textit{training-free} diffusion model compression framework designed to enhance the per-step inference efficiency of pre-trained diffusion models at both the input level and module level in a post-training manner. Our techniques are training-free and can thus be instantly applied to emerging diffusion models for rapid deployment, enabling a comparative study of their effectiveness versus reducing denoising steps. Specifically, at the input level, we propose a mixed-resolution denoising scheme that adopts a lower generation resolution in early denoising steps. As low‑frequency components dominate at this stage, a reduced resolution is sufficient. We find that this reduced resolution in early denoising steps can enhance low-frequency components, thereby improving final generation fidelity and achieving a win-win in both efficiency and fidelity. At the module level, we build on existing module caching techniques, which reuse module outputs across steps, and enhance them with a hybrid caching strategy as another dimension for redundancy reduction.
Our contributions are summarized as follows:

\begin{itemize}
    \item Our work, for the first time, studies a critical question for efficient diffusion model deployment: Under a post-training setting without fine-tuning, is it more effective to optimize for fewer denoising steps or to reduce the per-step inference cost? The answer to this question could be useful for both researchers and practitioners.
    
    \item To answer the above question, we propose PostDiff, a training-free framework for reducing diffusion model complexity at 
    both 
    the input and module levels, resulting in lower per-step inference cost. This framework enables a fair study of the relative effectiveness between fewer denoising steps and reduced per-step inference cost toward compute-optimal diffusion model deployment under a post-training setting.

    \item At the input level, we propose a mixed-resolution denoising scheme, which leverages the observation that reduced generation resolution in early denoising steps can enhance low-frequency components and final output quality, achieving a win-win in both accuracy and efficiency. In addition, to further reduce the overall computational cost at the module level, PostDiff employs a hybrid module caching method that strategically reuses the outputs of different modules on varying schedules across steps.

    \item Extensive evaluations show that PostDiff can advance the fidelity-efficiency frontier of state-of-the-art (SOTA) diffusion models, as shown in Fig.~\ref{fig:sota}, and provide valuable insights into best practices for efficient diffusion model deployment. For example, we find that reducing per-step inference cost is often more effective than decreasing the number of denoising steps to improve efficiency when aiming to preserve high generation fidelity.
    
\end{itemize}

In the following sections, we illustrate and analyze the effectiveness of each technique in PostDiff in Sec.~\ref{sec:resolution}, and finally, we combine and apply them to SOTA diffusion models in Sec.~\ref{sec:alltech} to answer our research question.

\section{PostDiff: The Proposed Framework}
\label{sec:resolution}

In this section, we first introduce the proposed mixed-resolution denoising strategy, which reduces per-step inference cost at the input level. We begin by analyzing the motivations from previous works in Sec.~\ref{sec:resolution_motivation}, followed by an illustration of our method, an analysis of its effectiveness, and its application to examine our key question in Sec.~\ref{sec:resolution_method}, Sec.~\ref{sec:resolution_visual}, and Sec.~\ref{sec:resolution_exp}, respectively. Finally, we describe how we employ and enhance module caching to improve efficiency at the module level in Sec.~\ref{sec:cache_method}.

\subsection{Motivations from Previous Works}
\label{sec:resolution_motivation}

\textbf{Low-frequency components dominate the early denoising steps.} 
Previous works~\cite{liu2024faster} find that the typical denoising process for image generation progresses through two phases: the semantics-planning phase and the fidelity-improving phase. The early semantics-planning phase focuses more on the layout and low-frequency information, starting from random noise and gradually establishing the basic structure. This indicates that a low generation resolution could be sufficient in the early denoising steps. As generation proceeds, more details, described by high-frequency information, are incrementally added, thus requiring a higher resolution.

This insight is also partially echoed and leveraged by previous works. For example,~\cite{teng2023relay} trains different diffusion models for early low-resolution denoising and later high-resolution denoising processes to avoid the propagation of inaccurate high-frequency components from early denoising steps to later steps. Additionally, cascade diffusion models \cite{ho2022cascaded,wang2023lavie} sequentially apply different diffusion models at different denoising stages with varying generation resolutions, finding that incorporating low-frequency information from low-resolution images during high-resolution stages can enhance generation quality. Nevertheless, all these methods require multiple diffusion models, introducing significant storage and training demands.

\textbf{Input resolution is a promising dimension for reducing inference cost.} High-resolution generation is a recent trend, as it provides images with more vivid details and accommodates greater information density \cite{zhang2025hidiffusion}. For example, SDXL \cite{podell2023sdxl} generates images at 1024x1024 resolution, while PixArt-$\Sigma$ \cite{chen2024pixartsigma} expands resolution capability to 4K. However, as resolution increases, computational demands rise dramatically, leading to significant latency \cite{Li2024Distri}. As such, recent works aim to reduce the computational load by decreasing the number of tokens in samples, using methods like token merging \cite{bolya2023token}, token pruning \cite{wang2024attention}, or token downsampling \cite{smith2024todo}, which effectively accelerate inference.

\begin{figure}
    \centering
    \includegraphics[width=1\linewidth]{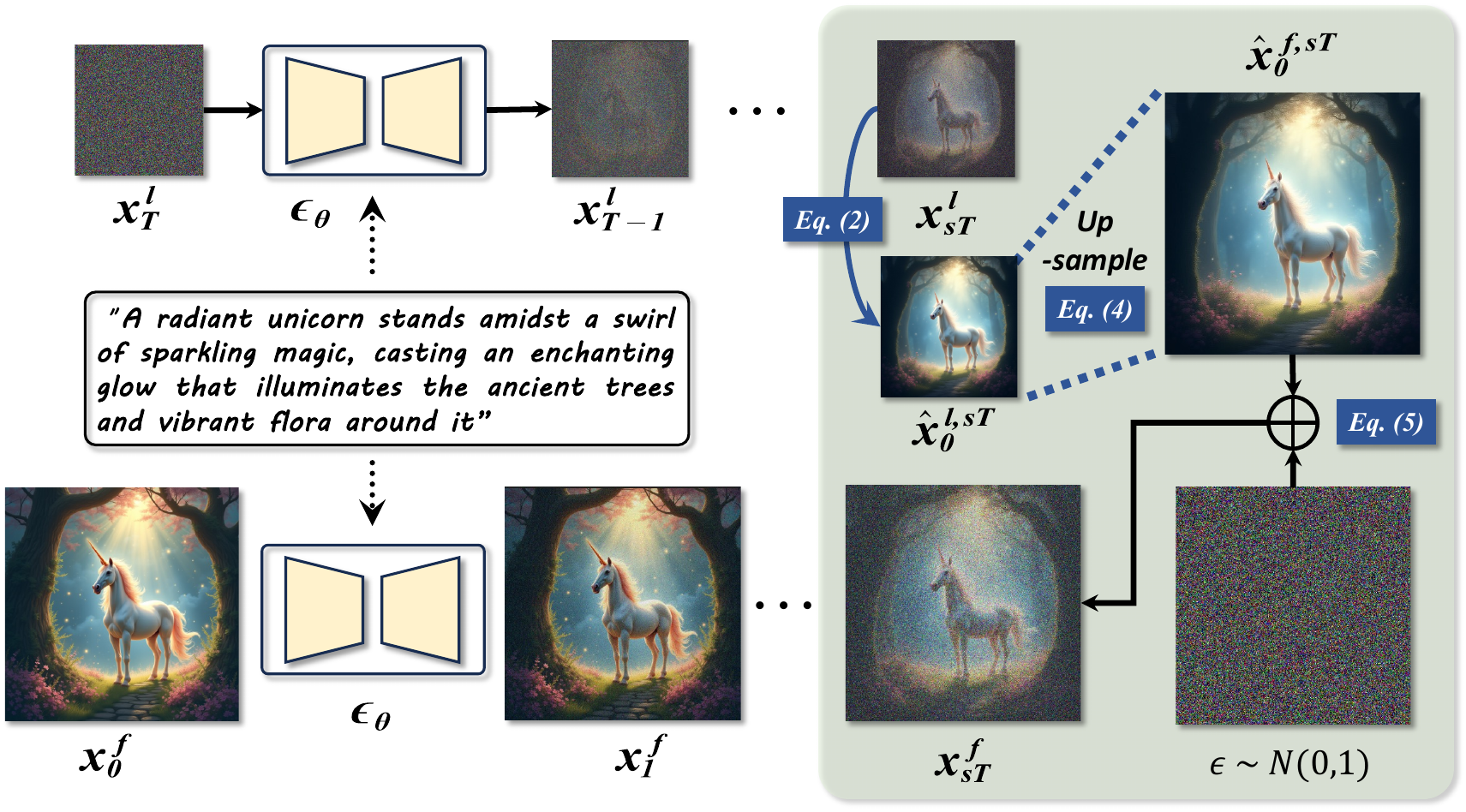}
    \vspace{-1em}
    \caption{An overview of our mixed-resolution denoising strategy.}
    \label{fig:overview}
    \vspace{-1em}
\end{figure}

\begin{figure*}
    \centering
    \includegraphics[width=0.99\linewidth]{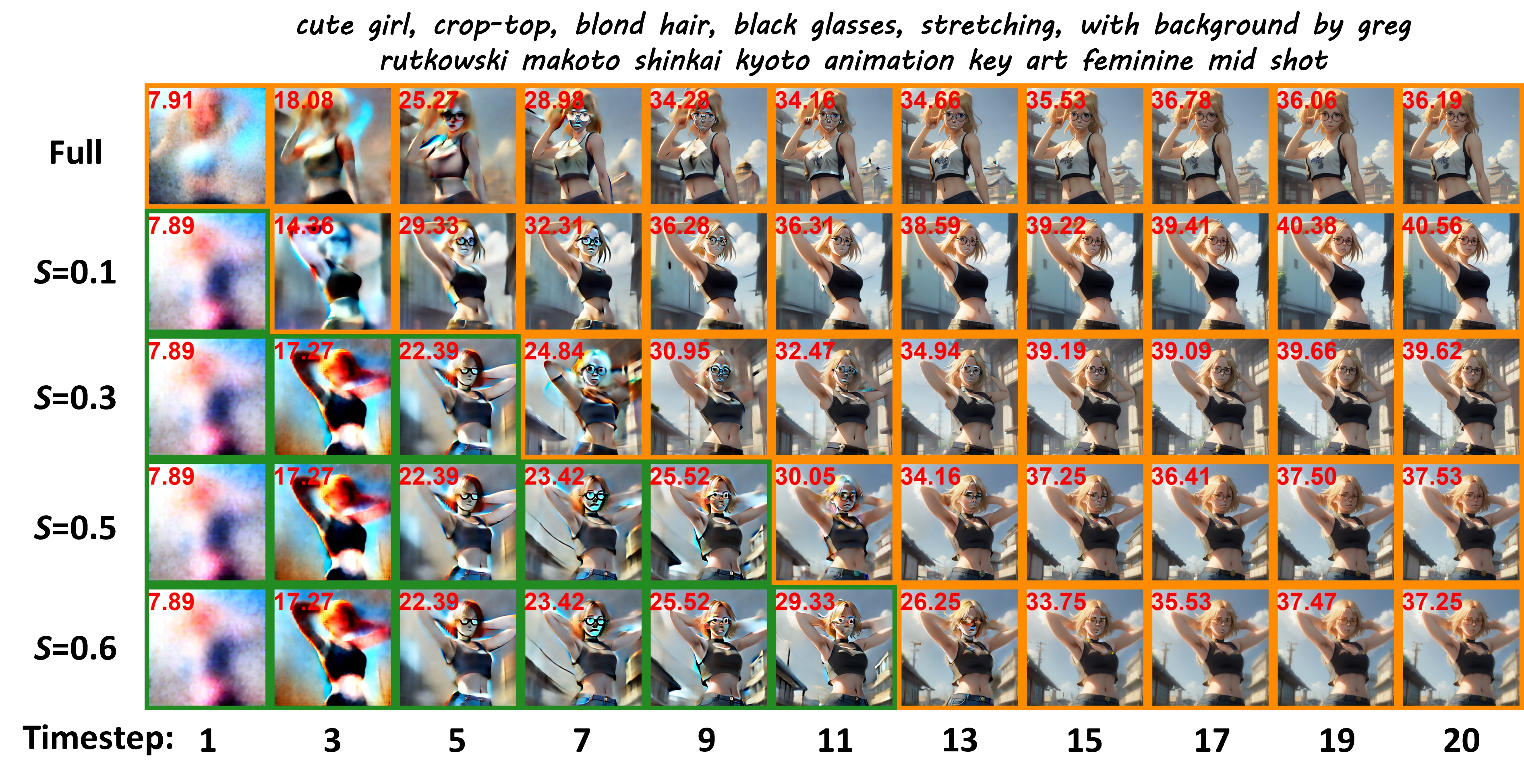}
    \vspace{-1em}
    \caption{Visualize the Clip Score after each denoising step using full resolution (row 1) and mixed resolution (rows 2-5). The steps using low generation resolution are highlighted in green and those using high generation resolution are highlighted in orange.}
    \label{fig:evolution}
    \vspace{-1em}
\end{figure*}

\subsection{Mixed-Resolution Denoising: Methodology}
\label{sec:resolution_method}

Motivated by the tendency for low-frequency components in early denoising steps and the effectiveness of input resolution as a design knob for efficiency,
we propose a simple yet effective post-training strategy: As shown in Fig.~\ref{fig:overview}, we adopt a low generation resolution in the early denoising steps to generate low-frequency structures, and then refine high-frequency details with high generation resolution in the later denoising steps.

\textbf{Formulation.} To generate an image with shape $(W, H, 3)$, we begin with Gaussian noise in the shape $(w, h, 4)$, representing the latent noise image $x_T$ obtained via Variational Autoencoders \cite{razavi2019generating}. For simplicity, we omit the channel dimension in the remainder of this paper. The denoising process proceeds over $T$ steps, using samplers like DDIM \cite{song2020denoising} or DPM-solver \cite{lu2022dpm} to obtain the denoised latent at each step. Taking DDIM \cite{song2020denoising} as an example, we compute the latent representation $x$ at step $t-1$ as follows:

\vspace{-1em}
\begin{align}
x_{t-1} = \sqrt{\alpha_{t-1}}\hat{x}_0^t  + x_t, \\
\hat{x}_0^t =  \frac{x_t - \sqrt{1 - \alpha_t} \cdot \epsilon_\theta(x_t, t)}{\sqrt{\alpha_t}}, \\
x_t = \sqrt{1 - \alpha_{t-1}} \cdot \epsilon_\theta(x_t, t),
\end{align}
where $\epsilon_{\theta}$ denotes the model \cite{Ronneberger2015unet, peebles2023scalable} that predicts the added noise, and $\{\alpha_t\}$ is a set of coefficients controlling the forward noise-adding process \cite{ho2020denoising}.

In our proposed mixed-resolution denoising strategy, we first introduce a lower-resolution latent representation, $x_T^{l}$, with shape $(\beta w, \beta h)$, where $0<\beta<1$ is a hyper-parameter that scales the resolution down. This low-resolution starting point reduces the computational load of the initial denoising phase. Another hyperparameter, $s$, controls the transition from low to full resolution. After reaching the denoising step $t=sT$, we compute the forecasted $\hat{x}_0^{l,t}$ at step $t$ using Eq. (2). Then, we upscale this result to full resolution via:

\vspace{-1em}
\begin{align} \hat{x}_0^{f,t} = \text{Upsample}(\hat{x}_0^{l,t}), \end{align}
where $\text{Upsample}$ refers to bilinear interpolation. Using this upscaled version, we derive the full-resolution latent $x_t^f$ at step $t$ as follows:

\vspace{-1em}
\begin{align} 
x_t^f = \sqrt{\alpha_t} \hat{x}_0^{f,t} + \sqrt{1 - \alpha_t} \epsilon, 
\end{align}
where $\epsilon$ is Gaussian noise with shape $(w, h)$. The remainder of the denoising process follows the original procedure. 

Note that we adopt this simple binary resolution strategy without bells and whistles to effectively demonstrate its concept and simplify the involved hyperparameters, thereby easing the use of our method in real-world deployment. More complex resolution schedules could potentially be proposed to further enhance fidelity-efficiency trade-offs.

\subsection{Visualization: How Low-Resolution Denoising Contributes to Boosted Final Performance}
\label{sec:resolution_visual}

To study the impact of our mixed-resolution denoising strategy on final performance and understand why it works, we visualize the generation quality after each denoising step, both with and without our method. Specifically, we calculate the Clip Score~\cite{hessel2021clipscore} for \(\hat{x}_0\) predicted at each denoising step and annotate it in the case study of Fig.~\ref{fig:evolution}, where different rows correspond to the original full-resolution denoising process as well as our method with various choices of $s$.

\textbf{Observations and analysis.}
We observe that \underline{(1)} with a moderate amount of low-resolution generation steps, the Clip Score of the final generated image can be improved, achieving a win-win in generation fidelity and efficiency; \underline{(2)} Regarding the evolution of the Clip Score across denoising steps, starting with a lower generation resolution results in lower Clip Scores initially, but serves as a better starting point: after switching to high generation resolution, the Clip Score increases more rapidly and gradually surpasses that of full-resolution generation. This echoes the analysis in Sec.~\ref{sec:resolution_motivation}, where reduced generation resolution in early denoising steps contributes to improved low-frequency components, thereby enhancing final generation quality; \underline{(3)} The balance between low-resolution and high-resolution steps is crucial, as both low-frequency structure and high-frequency details need to be preserved to ensure high final generation quality in the last denoising step.

This phenomenon demonstrates the underlying effectiveness of our mixed-resolution denoising strategy and is consistently observed across samples, as shown in the comprehensive evaluation in Sec.~\ref{sec:resolution_exp}.

\subsection{Mixed-Resolution Denoising: Evaluation}
\label{sec:resolution_exp}

\begin{figure}
    \centering
    \includegraphics[width=\linewidth]{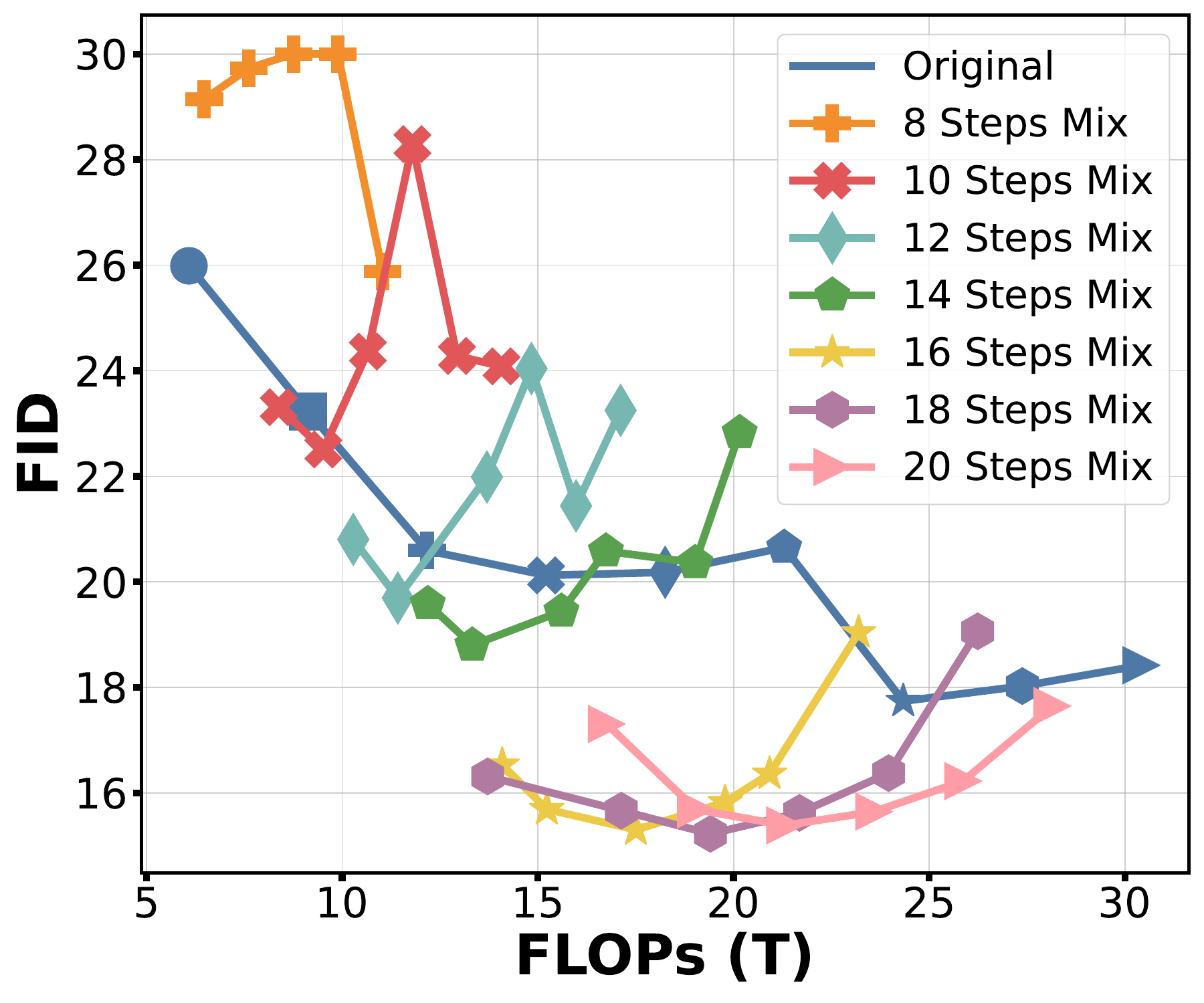}
    \vspace{-1.5em}
    \caption{The FID-FLOPs trade-off achieved under different denoising steps and varied mixed-resolution settings. The data points with the same number of steps are annotated using the same shape.}
    \label{fig:exp_mix_res}
    \vspace{-1.5em}
\end{figure}

We evaluate the effectiveness of our mixed-resolution denoising strategy and examine which is more effective: reduced input-level redundancy or fewer denoising steps.

\textbf{Setup.} We use Stable Diffusion (SD) V1.5 (DreamShaper-7 version)~\cite{rombach2022high}, a popular latent diffusion model (LDM), as our model backbone and evaluate text-to-image generation performance on the MS-COCO 2014 validation dataset~\cite{lin2014microsoft}. We vary the total denoising steps \( T = \{8, 10, 12, 14, 16, 18, 20\} \) and apply our mixed-resolution denoising to each denoising step setting with varied \( s = \{0.1, 0.2, 0.3, 0.4, 0.5, 0.6\} \). Without bells and whistles, we set $\beta=0.5$ as our low-resolution setting, i.e., half the full resolution, and ablation studies on this are provided in the supplementary material.

\textbf{Observations and analysis.}
As shown in Fig.~\ref{fig:exp_mix_res}, we observe that \underline{(1)} mixed-resolution denoising can boost achievable generation fidelity by notably reducing the optimal FID (lower is better) while simultaneously improving generation efficiency, aligning with the analysis of the role of low resolution in early denoising steps in Sec.~\ref{sec:resolution_motivation} and Sec.~\ref{sec:resolution_visual}; \underline{(2)} With more inference steps (\( T \geq 14 \)), the optimal values of \( s \) are around 0.4$\sim$0.5, offering the best FID-FLOPs trade-off. For example, adopting \( s = 0.5 \) with 20 denoising steps can save about 37\% FLOPs while achieving an FID reduction of approximately 2 compared to the original full-resolution model.
\underline{(3)} When aiming for relatively high generation fidelity, e.g., a $<$20 FID, which aligns with real-world application demands, maintaining a high number of denoising steps and reducing per-step inference cost via mixed resolution is a more cost-effective strategy than reducing denoising steps. Conversely, when targeting extreme efficiency, e.g., reducing $>$60\% FLOPs, using fewer denoising steps becomes increasingly beneficial, as the effectiveness of mixed-resolution denoising diminishes in low-step generation with larger gaps between steps.

\textbf{Calibration for best hyperparameters.} In practice, to rapidly determine the best transition step $s$ (as well as the total number of denoising steps) for a given diffusion model provided by the users, we can use a small calibration set to obtain the optimal setting. We show in the supplementary materials that performance on a small calibration set is highly correlated with that on the full evaluation set.

\subsection{The Hybrid Module Caching Strategy}
\label{sec:cache_method}

In addition to the resolution dimension, we further reduce model complexity at the module level by integrating existing module caching techniques~\cite{ma2024deepcache, wimbauer2024cache} with an enhanced hybrid caching strategy.

\textbf{Motivation.}
Previous works~\cite{ma2024deepcache, wimbauer2024cache} have found that feature maps remain highly similar across all blocks throughout the entire denoising process, indicating the potential effectiveness of caching and reusing across steps. We have further validated this through cross-step feature map distance visualization in the supplementary materials.  
Considering that (1) the text guidance provided by cross-attention layers primarily determines the image layout, i.e., the low-frequency structure of the image, and (2) the semantics-planning phase for layout generation occurs in the early denoising stages~\cite{liu2024faster}, as also discussed in Sec.~\ref{sec:resolution_motivation}, it is feasible to use the cached cross-attention feature map from early denoising steps, generated at a low resolution with our mixed-resolution denoising, for the later denoising phase, as the low-frequency information is determined early on.

\textbf{Our caching strategy.}
We employ a hybrid module caching strategy to maximize reuse opportunities in a pre-trained diffusion model. To achieve this, we integrate the following two caching mechanisms: (1) similar to DeepCache~\cite{ma2024deepcache}, we cache the deep skip branches of the U-Net, a major component of a wide range of SOTA diffusion models~\cite{rombach2022high,podell2023sdxl,kim2023architectural,luo2023latent}, and reuse them for the subsequent \( k \) steps. Here, \( k \) controls how often the cache is updated, balancing efficiency gains and potential degradation, and is set to 2 throughout the evaluation; (2) We cache the cross-attention layers at a specific denoising step and reuse them in all subsequent steps \cite{liu2024faster}.

\textbf{Design choices for caching cross-attention.} A major role of cross-attention is to introduce text guidance into the generation process. This is often achieved through classifier-free guidance (CFG), which provides a controllable method for incorporating conditional guidance into image generation \cite{dhariwal2021diffusion, Ho2022ClassifierFreeDG, Nichol2021GLIDETP}. Specifically, the text guidance from the model's implicit classifier is derived as follows:  
\begin{align}
    \epsilon_{\theta}(x_t, t, w, c) = \epsilon_{\theta}(x_t, t, \emptyset) + w \left( \epsilon_{\theta}(x_t, t, c) - \epsilon_{\theta}(x_t, t, \emptyset) \right),
\end{align}  
where $\emptyset$ represents the null text condition. 

Considering that CFG may be redundant in the later stages of inference, as the layout has already been organized in the early stages \cite{castillo2023adaptive, liu2024faster}, we abandon CFG after a certain step \( m \). When combined with our mixed-resolution denoising scheme, most CFG operations are performed under a low generation resolution, further reducing computational costs.

Based on the type of guidance to be reused, there are four design choices for caching the cross-attention maps for subsequent steps:  

\begin{itemize}
    \item \textbf{Ave} \cite{liu2024faster}: \( CA_{\text{cache}} = \frac{1}{2} \left( CA_t^{c} + CA_t^{\emptyset} \right) \)  
    \item \textbf{Cond}: \( CA_{\text{cache}} = CA_t^{c} \)  
    \item \textbf{Uncond}: \( CA_{\text{cache}} = CA_t^{\emptyset} \)  
    \item \textbf{CFG}: \( CA_{\text{cache}} = CA_t^{\emptyset} + w(CA_t^{c} - CA_t^{\emptyset}) \)  
\end{itemize}  
Here, \( CA_t^{c} \) and \( CA_t^{\emptyset} \) are the cross-attention  conditioned on the input text and null text, respectively. Using either of these maps or a mixture of them results in varying degrees of conditioning in the subsequent generation process.

\begin{table}[t]  \centering
  \caption{Ablation study on different caching strategies ($m$, Design Choice).}
  \vspace{-0.5em}
  \addtolength{\tabcolsep}{3.0pt}
  \resizebox{0.45\textwidth}{!}{
      \begin{tabular}{cc|cc}
          \Xhline{1px}
          \textbf{Setting}  & \textbf{FLOPs} (T) $\downarrow$ & \textbf{FID} $\downarrow$ & \textbf{Clip Score} $\uparrow$ \\
          \hline
          Original  & 30.420 & 18.42  & 30.80  \\
          Original w/o CFG  & 15.210 & 31.98  & 25.52  \\
          DC  & 17.787 & 17.79  &   30.75  \\
          \hline
          DC+CA (5, Ave)  &\multirow{4}{*}{11.610} & 18.77  & 28.40     \\
          DC+CA (5, Cond)  & & 18.82  &  29.20   \\
          DC+CA (5, Uncond) & & 20.33  & 25.50     \\
          DC+CA (5, CFG)  & & 103.71  &  18.22    \\
          \hline
          DC+CA (10, Ave)  &\multirow{4}{*}{15.061} & 21.13  & 30.03   \\
          DC+CA (10, Cond)  & &  21.26 & 30.11  \\
          DC+CA (10, Uncond) & & 21.19  &  29.83  \\
          DC+CA (10, CFG)  & & 24.00  & 29.12    \\
          \hline
          DC+CA (15, Ave) &\multirow{4}{*}{16.360} & 21.70  & 30.35   \\
          DC+CA (15, Cond)  & & 21.67  &  30.37  \\
          DC+CA (15, Uncond) & & 21.82  & 30.34   \\
          DC+CA (15, CFG) &  &  19.43  & 30.48   \\
          \Xhline{1px}
      \end{tabular}  }   
        \vspace{-1em}
      \label{tab:cache_ablation}
\end{table}

\textbf{Exploring the best cross-attention caching strategy.} We empirically examine combinations of different caching strategies, including DeepCache (DC) combined with the four design choices for cross-attention caching on top of an SD V1.5 model. As shown in Tab.~\ref{tab:cache_ablation}, we observe that \underline{(1)} fully abandoning CFG leads to a collapse in performance, while incorporating the cross-attention caching mechanism can better balance performance and speed; \underline{(2)} adding CA on top of DC further reduces computations due to redundant CFG usage in later denoising phases. Intuitively, using a smaller \( m \), i.e., abandoning CFG in earlier timesteps, reduces more FLOPs at the cost of a slight decrease in generation quality; \underline{(3)} among the four design choices, ``Cond” generally performs best, slightly surpassing ``Ave” at larger \( m \) values. Therefore, we adopt “Cond” as the \( CA_{\text{cache}} \) choice in all subsequent experiments.

\section{PostDiff: Combine All Techniques}
\label{sec:alltech}

\subsection{Experiment Setup}

We apply our PostDiff framework, integrating mixed-resolution denoising and hybrid module caching, to several SOTA diffusion models, including SD V1.5 (DreamShaper-7 version) \cite{rombach2022high}, Latent Consistency Model (LCM) \cite{luo2023latent, song2020denoising}, Stable Diffusion XL (SDXL) \cite{podell2023sdxl}, and PixArt-$\alpha$ \cite{chen2023pixart}. We evaluate text-to-image generation performance on the MS-COCO 2014 validation dataset~\cite{lin2014microsoft}. PostDiff settings are calibrated on a small set for each model, with the detailed configurations outlined as follows:

\begin{enumerate}
    \item \textbf{SD V1.5} \cite{rombach2022high}: We use the DreamShaper-v7 fine-tuned version with a final resolution of 768$\times$768. Settings: $\beta=1/2$, $s=1/2$, $w=7.5$, $m=15$.
    
     \item \textbf{Latent Consistency Model (LCM)} \cite{luo2023latent, song2020denoising}: A distilled version of Stable Diffusion designed for few-step generation, producing 768$\times$768 images. Settings: $s=1/2$, $\beta=1/2$, $w=7.5$, $m=4$. Note that since CFG is integrated into a guidance embedding \cite{meng2023distillation} and the model has a few-step nature, only $CA_{\text{cache}}$ is used without DC.
     
    \item \textbf{Stable Diffusion XL (SDXL)} \cite{podell2023sdxl}: A large-scale diffusion model capable of generating 1024$\times$1024 images. Settings: $\beta=3/4$,  $s=1/5$, $w=5.0$, $m=15$.

    \item \textbf{PixArt-$\alpha$} \cite{chen2023pixart}: A transformer-based diffusion model generating 1024$\times$1024 images. Due to the absence of bypass connections, only $CA_{\text{cache}}$ is used without enabling DC. Settings: $\beta=3/4$, $s=1/2$, $w=4.5$, $m=15$.
\end{enumerate}

\subsection{Fewer Denoising Steps vs. Cheaper Per-step Inference}
\label{subsec:trade-off}

After combining the hybrid module caching with our mixed-resolution strategy to establish the complete PostDiff framework, we examine our key question by benchmarking with the FID-FLOPs/Clip Score-FLOPs trade-offs achieved by reducing denoising steps. Specifically, for SD V1.5, we use 20 denoising steps, a fixed mixed-resolution setting of \( s=1/2 \) and \( \beta=1/2 \), and vary cache settings \( m \) to create the trade-off. As for PixArt-\(\alpha\), we use a fixed 20 denoising steps with \( \beta = 3/4 \), and vary \( m \) and \( s \) to form the trade-off.

\begin{figure}[t]
    \centering
    \includegraphics[width=0.99\linewidth]{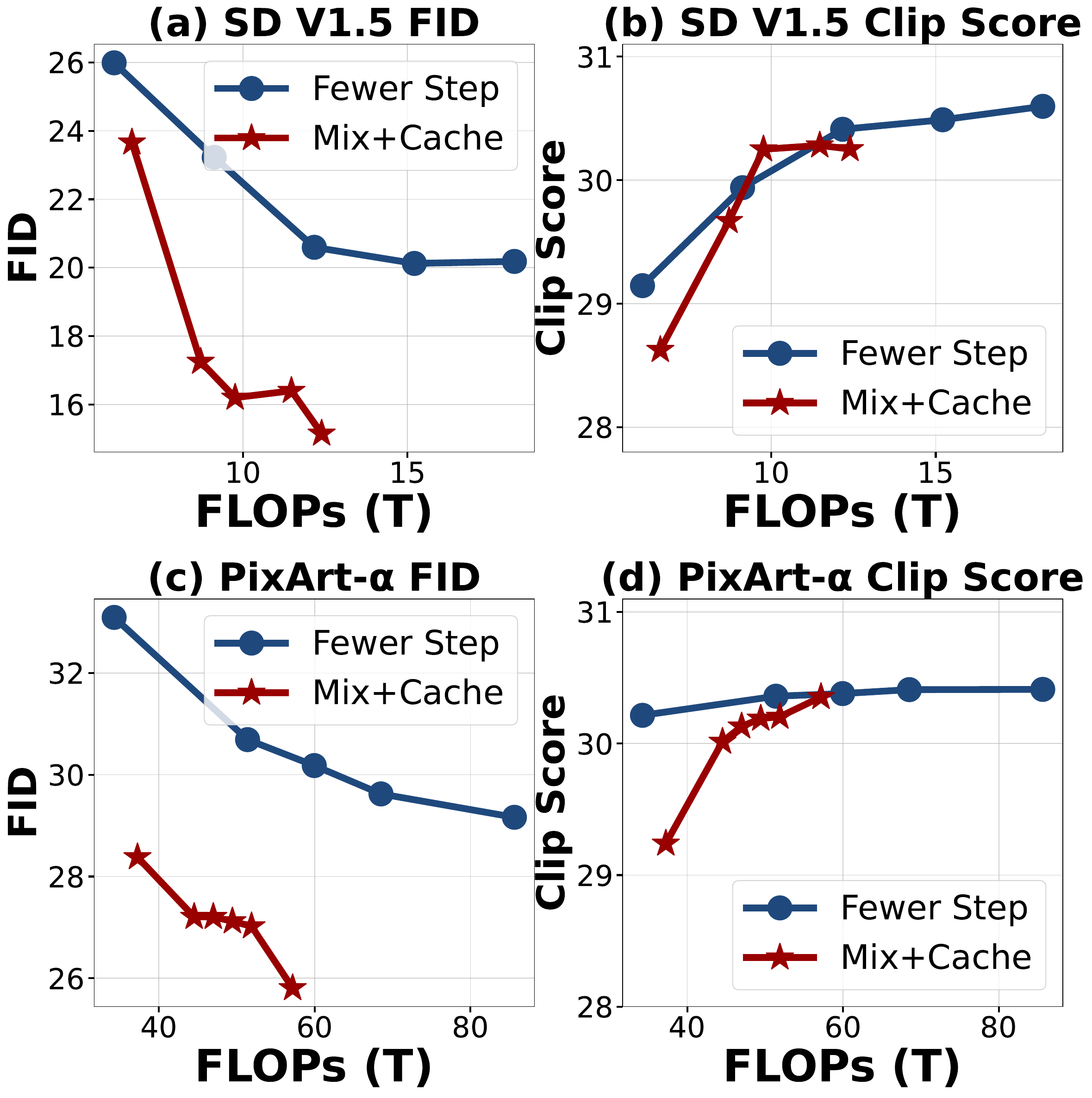}
    \caption{Comparison of FID-FLOPs/Clip Score-FLOPs trade-offs achieved by reducing per-step inference cost or the number of denoising steps on SD V1.5 and PixArt-\(\alpha\).}
    \label{fig:tradeoff_comparison}
    \vspace{-1em}
\end{figure}

\underline{Observations and insights.}
From Fig.~\ref{fig:tradeoff_comparison}, we observe that \underline{(1)} the complete PostDiff framework, incorporating both techniques, achieves the best FID-FLOPs trade-off across all FLOPs ranges. This demonstrates that exploiting model redundancy at both the input and module levels can yield better FID-FLOPs frontiers; \underline{(2)} in terms of semantic alignment measured by Clip Score, reducing per-step inference cost can achieve a comparable Clip Score-FLOPs trade-off compared to decreasing the number of denoising steps over a wide range; while in scenarios requiring extreme efficiency, decreasing the number of denoising steps can maintain better semantic alignment with a slightly higher Clip Score.  
Generally, reducing per-step inference cost is more effective than decreasing the number of denoising steps with better fidelity and comparable semantic alignment across a wide efficiency range.

\begin{table*}[h]  \centering
  \caption{Apply PostDiff with mixed-resolution denoising and hybrid module caching to SOTA diffusion models with varying numbers of denoising steps. Latency is measured as the single-image generation time on an NVIDIA A5000 GPU.}
  \addtolength{\tabcolsep}{3.0pt}
  \resizebox{0.95\textwidth}{!}{
      \begin{tabular}{c|ccc|ccccc}
          \Xhline{1px}
          \textbf{Model} &\textbf{Steps} &\textbf{Mix} &\textbf{Cache} & \textbf{FID} $\downarrow$ & \textbf{Clip Score} $\uparrow$ & \textbf{FLOPs} (T) $\downarrow$  & \textbf{Latency} (s) $\downarrow$  \\
           
          \hline
          \multirow{6}{*}{\textbf{SD V1.5} \cite{rombach2022high}}   & 8  &             &                          & 20.60   &  30.41           &   12.168          & 1.298\\
                                     & 16 &             &                          & 17.74  &  30.76           &   24.336          & 2.377\\
                                     & 20 &             &                          & 18.42  &  30.80           &   30.420         & 2.930\\
                                     &  \cellcolor{lightyellow}  20 & \cellcolor{lightyellow} \Checkmark  &     \cellcolor{lightyellow}                      & \cellcolor{lightyellow} 15.69  &  \cellcolor{lightyellow} 30.78           &   \cellcolor{lightyellow} 19.035           & \cellcolor{lightyellow} 1.945\\
                                     & \cellcolor{lightred}  20 &  \cellcolor{lightred} \Checkmark  &  \cellcolor{lightred} \Checkmark               &  \cellcolor{lightred} 16.65  &   \cellcolor{lightred}  30.25          &    \cellcolor{lightred} 11.129          & \cellcolor{lightred}  1.139\\
          \hline
          \multirow{5}{*}{\textbf{LCM} \cite{luo2023latent}} 
                                     & 4 &             &                          & 23.55  &    28.64         &  3.042           & 0.523\\
                                     & 8 &             &                          & 22.96  & 28.72            &  6.084           & 0.825 \\
                                     & \cellcolor{lightyellow} 8 & \cellcolor{lightyellow} \Checkmark  &  \cellcolor{lightyellow}                         & \cellcolor{lightyellow} 21.01  &  \cellcolor{lightyellow}  28.92          &   \cellcolor{lightyellow} 3.807          & \cellcolor{lightyellow} 0.666\\
                                     &  \cellcolor{lightred} 8 &  \cellcolor{lightred} \Checkmark  &  \cellcolor{lightred} \Checkmark              &  \cellcolor{lightred} 22.76  &   \cellcolor{lightred}  28.58          &    \cellcolor{lightred} 3.686          &  \cellcolor{lightred} 0.651\\
          \hline
         \multirow{6}{*}{\textbf{SDXL} \cite{podell2023sdxl}}   & 8  &             &             &   	18.01          & 30.92   &    47.856         &  2.843          \\
                                     & 16 &             &                          & 14.35  &   31.92	          &  95.712           & 5.211\\
                                     & 20 &             &                          & 14.10 &   31.95          &      119.641       & 6.521 \\
                                     & \cellcolor{lightyellow} 20 & \cellcolor{lightyellow} \Checkmark  &   \cellcolor{lightyellow}                        & \cellcolor{lightyellow} 14.37  &  \cellcolor{lightyellow}  31.62	          &  \cellcolor{lightyellow} 109.264           & \cellcolor{lightyellow} 5.950\\
                                     &  \cellcolor{lightred} 20 &  \cellcolor{lightred} \Checkmark  &  \cellcolor{lightred} \Checkmark               & \cellcolor{lightred}  14.18  & \cellcolor{lightred}  31.11           &   \cellcolor{lightred} 52.682          & \cellcolor{lightred}  3.119\\
                              
        \hline
         \multirow{6}{*}{\textbf{PixArt-$\alpha$} \cite{chen2023pixart}}   & 8  &             &             &   33.09          & 30.21	   & 34.256            & 3.031            \\
                                     & 16 &             &                          &  29.62 &  30.40 	          &  68.512           & 5.725\\
                                     & 20 &             &                          & 29.16 &   30.41	          &   85.640          & 7.093\\
                                     & \cellcolor{lightyellow} 20 & \cellcolor{lightyellow} \Checkmark  &   \cellcolor{lightyellow}                        & \cellcolor{lightyellow} 25.99  &  \cellcolor{lightyellow}  30.38	          &  \cellcolor{lightyellow} 66.990           & \cellcolor{lightyellow} 5.645\\
                                     &  \cellcolor{lightred} 20 &  \cellcolor{lightred} \Checkmark  &  \cellcolor{lightred} \Checkmark               &  \cellcolor{lightred} 25.44  &  \cellcolor{lightred}  30.23	           &     \cellcolor{lightred} 54.718         &  \cellcolor{lightred} 4.768\\
                                   
          \Xhline{1px}

      \end{tabular}  }   
      
      \label{tab:sota_compare}
\end{table*}

\subsection{Evaluation on SOTA Diffusion Models}
We further apply PostDiff to more diffusion models.
As shown in Tab.~\ref{tab:sota_compare}, we present the evaluation results in terms of both generation quality and efficiency under (1) reduced denoising steps or (2) maximal denoising steps with reduced model complexity via PostDiff.

\begin{figure*}
    \centering
    \includegraphics[width=0.97\linewidth]{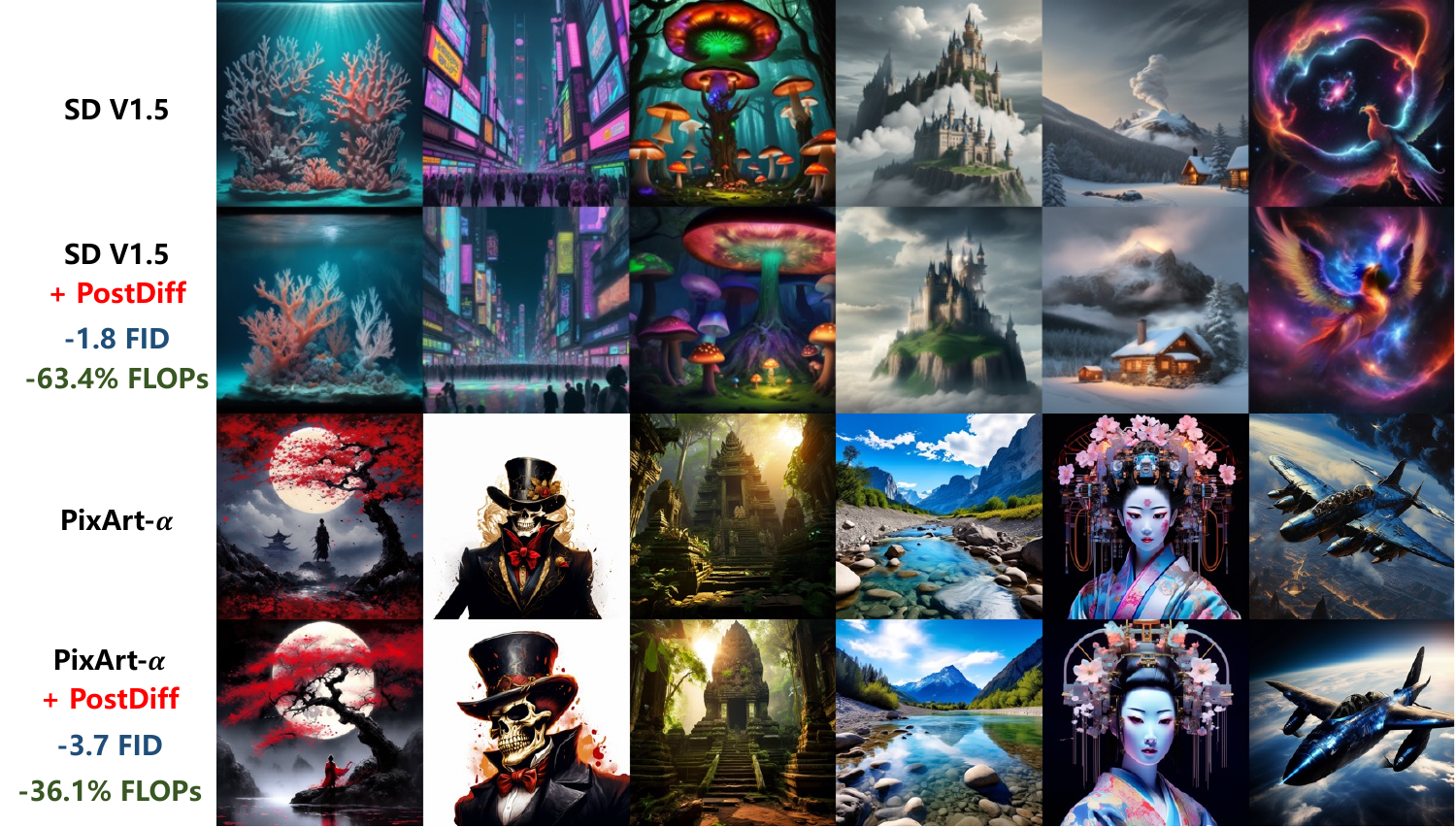}
    \caption{Visualize the generation results of different models w/o and w/ using our PostDiff.}
    \label{fig:visualization}
    \vspace{-1em}
\end{figure*}

\textbf{Observations and analysis.} We observe that:  
\underline{(1)} With reduced denoising steps, the improved efficiency comes at the cost of degraded generation quality, reflected in notably increased FID and reduced Clip score. In contrast, maintaining the maximum number of denoising steps while applying PostDiff better preserves generation quality, achieving lower (i.e., better) FID and comparable Clip scores compared to reducing the denoising steps. For instance, on the SD V1.5 model, PostDiff achieves a 63.14\% reduction in FLOPs while improving FID by 1.8 compared to the original model.
\underline{(2)} PostDiff consistently demonstrates effectiveness across various diffusion models, including both large-scale and small-scale ones, LDMs and LCM, as well as U-Net-based and transformer-based models. This effectiveness is due to the broad applicability of the insights leveraged in Sec.~\ref{sec:resolution_motivation} and Sec.~\ref{sec:cache_method}. We also visualize the generation results in Fig.~\ref{fig:visualization}.

\subsection{Benchmark with Other Training-Free Compression Methods}
\label{subsec:comparition_baselines}

We further benchmark PostDiff with other training-free diffusion model compression methods in Tab.~\ref{tab:benchmark_compression} and Tab.~\ref{tab:pixart}.

\textbf{Settings.}
For the U-Net-based model SD V1.5, we evaluate our method against DeepCache \cite{ma2024deepcache}, TGATE \cite{liu2024faster}, ToDo \cite{smith2024todo}, ToMe \cite{bolya2023token}, Faster Diffusion \cite{li2023faster}, and CA-ToMe \cite{saghatchian2025cached}. Latency is measured as the single-image generation time on an NVIDIA A5000 GPU. The cache interval for DeepCache is set to 2. For TGATE, we set \( m=5 \). For ToMe, we use a merging ratio of 0.3, while ToDo, Faster Diffusion, and CA-ToMe are evaluated using their default hyperparameters from their respective papers.  
For the transformer-based model PixArt-\(\alpha\), we compare PostDiff with TGATE \cite{liu2024faster} using \( m = 8 \), as well as DiTFastAttn \cite{yuan2024ditfastattn} and DuCa \cite{zou2024DuCa} with their default settings on an NVIDIA H100 GPU, since some methods encounter out-of-memory errors on the A5000.

\begin{table}[t]  \centering
  \caption{Benchmark with other compression methods on SD V1.5. }
  \addtolength{\tabcolsep}{3.0pt}
  \resizebox{0.49\textwidth}{!}{
      \begin{tabular}{c|cc|c}
          \toprule
          \textbf{Method}   & \textbf{FID} $\downarrow$ & \textbf{Clip Score} $\uparrow$ & \textbf{Latency} (s) $\downarrow$ \\
          \midrule
          Original  & 18.42  & 30.80  & 2.930 \\ \midrule
          DeepCache \cite{ma2024deepcache}  & 17.79  &   30.75  & 1.737\\
          TGATE  \cite{liu2024faster} & 19.51  &   29.55  &  1.992 \\
           ToDo \cite{smith2024todo}  & 18.28  & 30.75 & 2.520 \\
          ToMe   \cite{bolya2023token} & 17.43  &  30.55   &  2.730 \\
          Faster Diffusion \cite{li2023faster} & 17.90  &   30.40  & 2.005   \\
          CA-ToMe \cite{saghatchian2025cached} & 20.49 & 30.41 & 2.112 \\
         
          \midrule
          Ours  & 16.65  &   30.25  & 1.139 \\ 
          \bottomrule
      \end{tabular}  }   
      
      \label{tab:benchmark_compression}
      \vspace{-1em}
\end{table}

\begin{table}[t]
  \centering
  \caption{Benchmark with other training-free compression methods on PixArt-$\alpha$.}
  \resizebox{\columnwidth}{!}{%
  \begin{threeparttable}
   \begin{tabular}{c|cc|c}
          \toprule
          \textbf{Method}   & \textbf{FID} $\downarrow$ & \textbf{Clip Score} $\uparrow$ & \textbf{Latency} (s) $\downarrow$ \\
          \midrule
          Original  & 29.16  & 30.41  & 1.752 \\ \midrule
          TGATE  \cite{liu2024faster} & 	29.54  & 29.59    & 1.275 \\
          DiTFastAttn \cite{yuan2024ditfastattn} & 45.46 & 30.13    & 1.103\tnote{$\dagger$ }  \\
          DuCa  \cite{zou2024DuCa}            & 28.08 &30.12  & 1.724 \\  
          \midrule
          Ours   & 25.44 & 30.23  & 1.382	\\ 
          \bottomrule
      \end{tabular}
\begin{tablenotes}
  \footnotesize
  \item[$\dagger$ ] Faster speed is achieved with FlashAttention \cite{dao2022flashattention}.
  \end{tablenotes}
\end{threeparttable}
  }
\vspace{-1em}
  \label{tab:pixart}
\end{table}

\textbf{Observations.}
Compared to previous methods that accelerate the generation process either at the module level (e.g., DeepCache, TGATE, DuCa) by reusing computations across denoising steps or at the input level (e.g., ToMe, ToDo) by compressing redundant image tokens, our method uniquely integrates both approaches, resulting in a greater reduction in inference cost. Additionally, leveraging low-resolution images in the early stages enhances the denoising process, allowing our method to achieve the lowest FID with a competitive Clip Score across both SD V1.5 and PixArt-$\alpha$ models.

These experiments highlight the benefits of leveraging redundancy at both the input and module levels. Integrating the aforementioned training-free compression techniques into PostDiff could further enhance efficiency, which we leave for future work.

\section{Related Work}
\label{sec:related-work}

\textbf{Diffusion models.}
Diffusion models have emerged as a powerful approach for generative tasks, modeling the generative process as a denoising sequence to enable high-quality image generation \cite{ho2020denoising, song2020denoising}. Latent Diffusion Models (LDMs) \cite{rombach2022high} leverage Variational Autoencoders (VAEs) \cite{razavi2019generating} to operate in a compressed latent space, reducing memory demands and enabling efficient high-resolution generation. Open-source models like Stable Diffusion V1.5 \cite{rombach2022high}, Stable Diffusion XL \cite{podell2023sdxl}, and PixArt \cite{chen2023pixart, chen2024pixartsigma} have further expanded accessibility and are now applied across multiple domains \cite{gao2023implicit, du2024towards, guo2024context}. Despite their impressive generative quality, diffusion models often suffer from long generation times, limiting their broader applicability.

\textbf{Few-Step diffusion models.}
A primary focus in accelerating diffusion models is reducing the number of timesteps. 
DDIM \cite{song2020denoising} modifies DDPM \cite{ho2020denoising} by introducing non-Markovian sampling, enabling faster generation. DPM-Solver \cite{lu2022dpm} applies exponential integrators to approximate solutions to diffusion ODEs, enhancing sampling efficiency. Other approaches, such as consistency models \cite{luo2023latent, song2023consistency}, aim to map points on the ODE trajectory to the original sample, effectively reducing inference steps. Adversarial training is also employed to enable one-step or few-step diffusion \cite{yin2024improved, yin2024onestep, xu2024ufogen}. 
However, these few-step models still incur non-trivial training/distillation cost.

\textbf{Efficient diffusion models.}
Efforts to accelerate diffusion models extend from data optimization to model optimization. From the data perspective, prior works have reduced computational requirements by merging similar tokens \cite{bolya2023token} or pruning redundant tokens \cite{wang2024attention}. From the model perspective, recent works focus on identifying redundant feature maps across timesteps and using caching mechanisms to reuse these maps, significantly reducing computational costs \cite{ma2024deepcache, wimbauer2024cache}. Additional approaches aim to slim down models through quantization to lower bit precision \cite{he2023ptqd, shang2023posttrainingquantizationdiffusionmodels, zhao2024viditqefficientaccuratequantization} or by pruning unnecessary parameters \cite{castells2024ld, NEURIPS2023_35c1d69d, whalen2025earlybird}. Additionally, several systems have been developed specifically to accelerate diffusion model inference \cite{Li2024Distri, kodaira2023streamdiffusion}. 
\section{Conclusion}
\label{sec:conclusion}

In this work, we explore a critical question for compute-optimal diffusion model deployment: under a post-training setting, is it more effective to reduce denoising steps or decrease per-step inference cost? To investigate this, we propose PostDiff, a training-free framework that enhances diffusion model efficiency while systematically examining this trade-off.  
PostDiff reduces redundancy at two levels: at the input level, through a mixed-resolution denoising scheme that enhances low-frequency components in early steps, and at the module level, with a hybrid caching strategy.  
Extensive evaluations demonstrate that reducing per-step inference cost often achieves a better efficiency-fidelity balance than reducing denoising steps across a wide efficiency range. PostDiff not only offers a new perspective on the fidelity-efficiency trade-off for diffusion models but also provides practical insights for their deployment on resource-constrained platforms.

\section*{Acknowledgment}

This work is supported by the National Science Foundation (NSF) through CCRI funding (Award Number: 2016727), RTML funding (Award Number: 1937592), and an NSF CAREER award (Award Number: 2048183), as well as by CoCoSys, one of the seven centers in JUMP 2.0, a Semiconductor Research Corporation (SRC) program sponsored by DARPA.

{
    \small
    \bibliographystyle{ieeenat_fullname}
    \bibliography{ref}
}
\clearpage
\appendix
\setcounter{page}{1}
\maketitlesupplementary

In this supplementary material, we provide additional experimental results, analyses, and visualizations to complement our main paper. 

\begin{itemize}
    \item We provide an analysis on the module-level redundancy across denoising steps in \cref{sec:cache_motivation}.
    \item We examine the impact of varying the resolution scale \( \beta \) used in our mixed-resolution denoising on the final performance in \cref{sec:beta}.
    \item We show the ablation studies of hyperparameters $m$ and $k$ in \cref{sec:ablation}.
    \item We demonstrate the effectiveness of using a small calibration set to determine the optimal hyperparameters for mixed-resolution denoising in \cref{sec:cal}.
    \item We further evaluate the effectiveness of the proposed PostDiff on Imagenet21K\_Recaption dataset in \cref{sec:imagenet}.
    \item We apply fine-tuning on top of PostDiff to further improve its performance in \cref{sec:finetune}.
    \item We provide an overview figure to illustrate our hybrid module caching strategy in \cref{sec:caching}.
    \item We compare our PostDiff with prior works in \cref{sec:compare}.
    \item We present additional visual examples and their corresponding prompts in \cref{sec:visual} and \cref{sec:prompt}, respectively.
\end{itemize}

\section{Motivations for Module-level Redundancy from Profiling and Previous Works}
\label{sec:cache_motivation}

\begin{figure}
    \centering
    \includegraphics[width=1\linewidth]{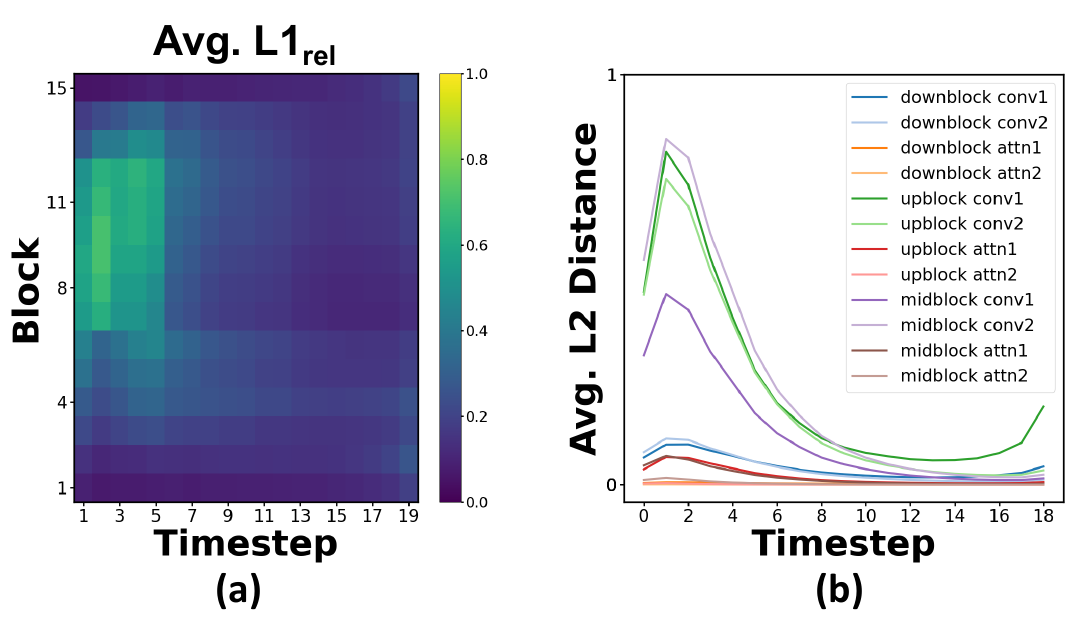}
    \vspace{-2em}
    \caption{Visualize the feature map distance across consecutive denoising steps based on (a) block index, and (b) block types.}
    \label{fig:simi}
    \vspace{-1em}
\end{figure}

To profile the module-level redundancy across denoising steps, we calculate the average relative $L_1$ distances \cite{wimbauer2024cache} of the $i$-th block's feature map between two consecutive steps, i.e., $t$ and $t-1$, as 
$
    L1_{\text{rel}}(i, t) = \left\| F_i(x_t) - F_i(x_{t-1}) \right\|_1 / \left\| F_i(x_t) \right\|_1
$,
where $F_i(\cdot)$ represents the output feature map of layer $i$. As shown in Fig.~\ref{fig:simi} (a), the feature maps are highly similar across all blocks throughout the entire denoising process, indicating the potential effectiveness of caching and reuse across steps. This insight is utilized by caching-based compression methods~\cite{ma2024deepcache, wimbauer2024cache}.

Delving deeper, we calculate the average \( L_2 \) distances of different types of layers' feature maps between two consecutive timesteps, as shown in Fig.~\ref{fig:simi} (b). We observe that attention layers' feature maps have relatively low distances across all steps compared to convolution layers. This stability enables effective caching with minimal impact on image quality. Additionally, considering that (1) the text guidance provided by cross-attention layers primarily determines the image layout, i.e., the low-frequency structure of the image, and (2) the semantics-planning phase for layout generation occurs in the early denoising stages~\cite{liu2024faster}, as also discussed in Sec. 2.1 of our main paper, it is feasible to use the cached cross-attention feature map in the later denoising phase, as the low-frequency information has been determined early on.

\section{Impact of Varying the Resolution Scale $ \beta $}
\label{sec:beta}
We analyze the effect of varying \( \beta \), the scaling factor applied to the resolution discussed in Sec. 2.2 of our main paper, on the performance of the mixed-resolution strategy.

\begin{figure}
    \centering
    \includegraphics[width=0.9\linewidth]{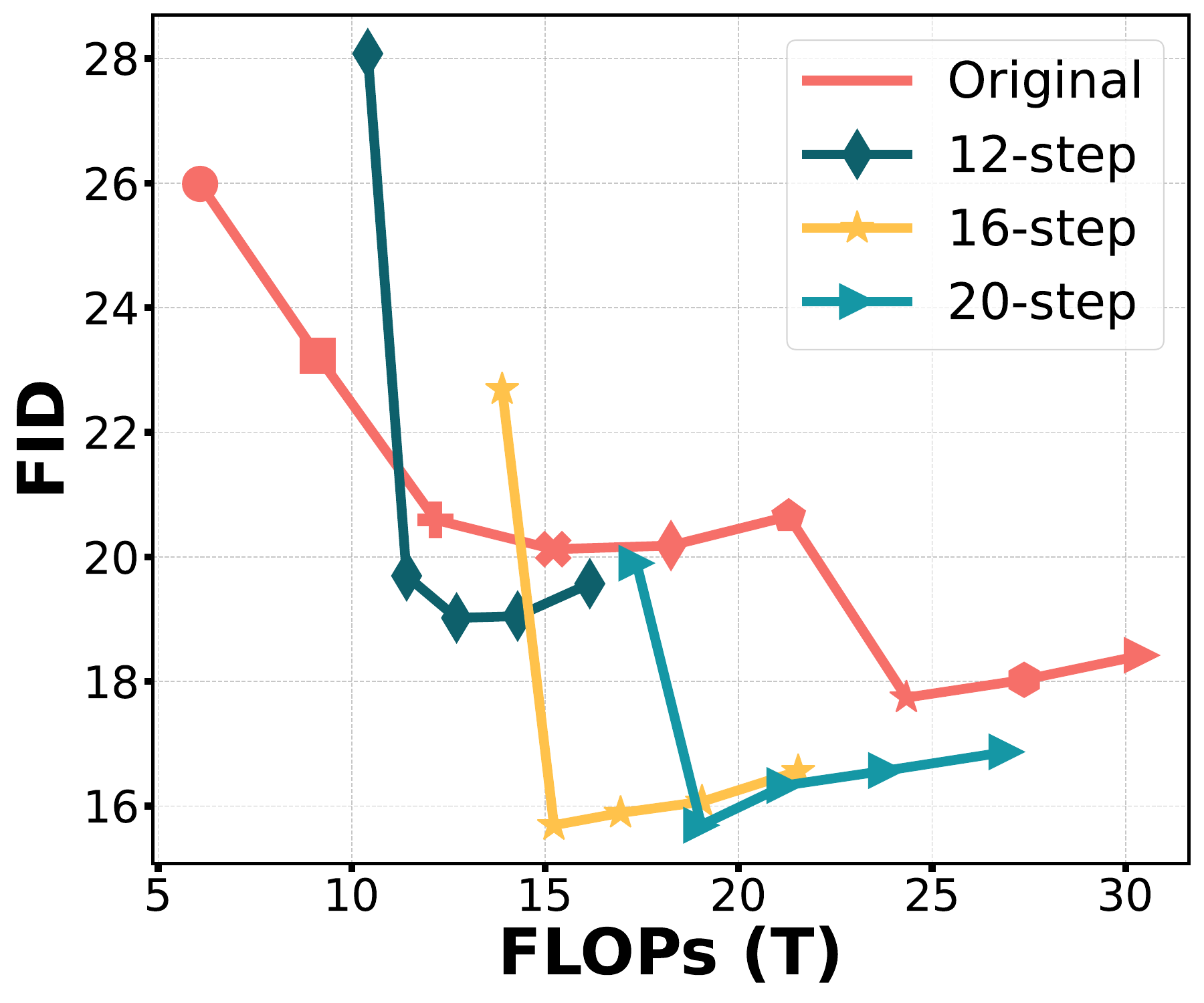}
    \caption{The FID-FLOPs trade-off achieved under different denoising steps and different $\beta$. The data points with the same number of denoising steps are annotated using the same shape.}
    \label{fig:multiple_beta}
\end{figure}

\textbf{Setup.} We use SD V1.5 (DreamShaper-7 version) as our backbone and the MS-COCO 2014 validation dataset as the test set. The scale factor \( \beta = \{0.375, 0.5, 0.625, 0.75, 0.875\} \) (from the leftmost points to the rightmost points on each curve in \cref{fig:multiple_beta}) is varied while keeping \( s = 0.5 \) fixed, i.e., low generation resolution is applied only in the first half of the denoising steps.

\textbf{Observations.} As illustrated in \cref{fig:multiple_beta}, we observe that \underline{(1)} $\beta = 0.5$ is generally a close-to-optimal choice, offering relatively low FIDs while significantly reducing FLOPs. As such, we adopt this design choice in our main paper; \underline{(2)} as $\beta$ increases from 0.5 to 0.875, the FIDs increase slightly. This is likely due to the low-resolution part becoming closer to the full resolution, which introduces more inaccurate high-frequency components during the early denoising stage, as analyzed in Sec. 2.1 of our main paper; \underline{(3)} smaller $\beta$ values result in lower FLOPs, while an extremely small $\beta = 0.375$ leads to a notable FID increase, as the extremely low resolution makes it difficult to recover fine details.

\section{Ablation Studies on $k$ and $m$} 
\label{sec:ablation}
We conducted ablation studies on $k$ (cache update frequency) and $m$ (the point at which CFG is abandoned) using SD V1.5 on an NVIDIA A5000 GPU. As shown in \cref{tab:ablation}, a larger $k$ (i.e., more cache reuse) or a smaller $m$ (i.e., abandoning CFG earlier) results in greater degradation of generation quality. Based on this analysis, we adopt $k=2$ and $m=0.75$ in our paper, which strikes a sweet-spot efficiency-fidelity trade-off.

\begin{table}[h]  
  \centering
  \caption{Ablation studies on $k$ and $m$.}
  \addtolength{\tabcolsep}{3.0pt}
  \resizebox{0.49\textwidth}{!}{
      \begin{tabular}{c|cc|c}
          \toprule
          \textbf{Settings}   & \textbf{FID} $\downarrow$ & \textbf{Clip Score} $\uparrow$ & \textbf{Latency} (s) $\downarrow$ \\
          \midrule
          Original          & 18.42 & 30.80 & 2.93 \\ \midrule
          $k=2$             & 16.65 & 30.25 & 1.14 \\  
          $k=3$             & 26.96 & 27.68 & 0.97 \\
          $k=4$             & 39.22 & 26.16 & 0.94 \\
          $k=5$             & 51.39 & 25.84 & 0.75 \\ \midrule
          $m=0.45$          & 17.94 & 28.82 & 1.00 \\
          $m=0.6$           & 17.21 & 29.50 & 1.08 \\
          $m=0.75$          & 16.65 & 30.25 & 1.14 \\
          $m=0.9$           & 16.49 & 30.06 & 1.26 \\
          \bottomrule
      \end{tabular}  
  }   
  \label{tab:ablation}
\end{table}

\section{Effectiveness of Small Calibration Sets}
\label{sec:cal}

As mentioned in Sec. 2.4 of our manuscript, a small calibration set can be leveraged to determine the optimal hyperparameter \( s \), i.e., the portion of denoising steps performed at a low generation resolution. We validate the effectiveness of this strategy in this section.

\textbf{Setup.} Using SD V1.5 as the backbone, we randomly sampled subsets of the MS-COCO 2014 validation dataset with sizes of 500 and 1000. We tested the average Clip Scores of the generated images for \( s = \{0.0 \, (\text{Full}), 0.3, 0.5, 0.6\} \). 

\begin{figure}
    \centering
    \includegraphics[width=0.9\linewidth]{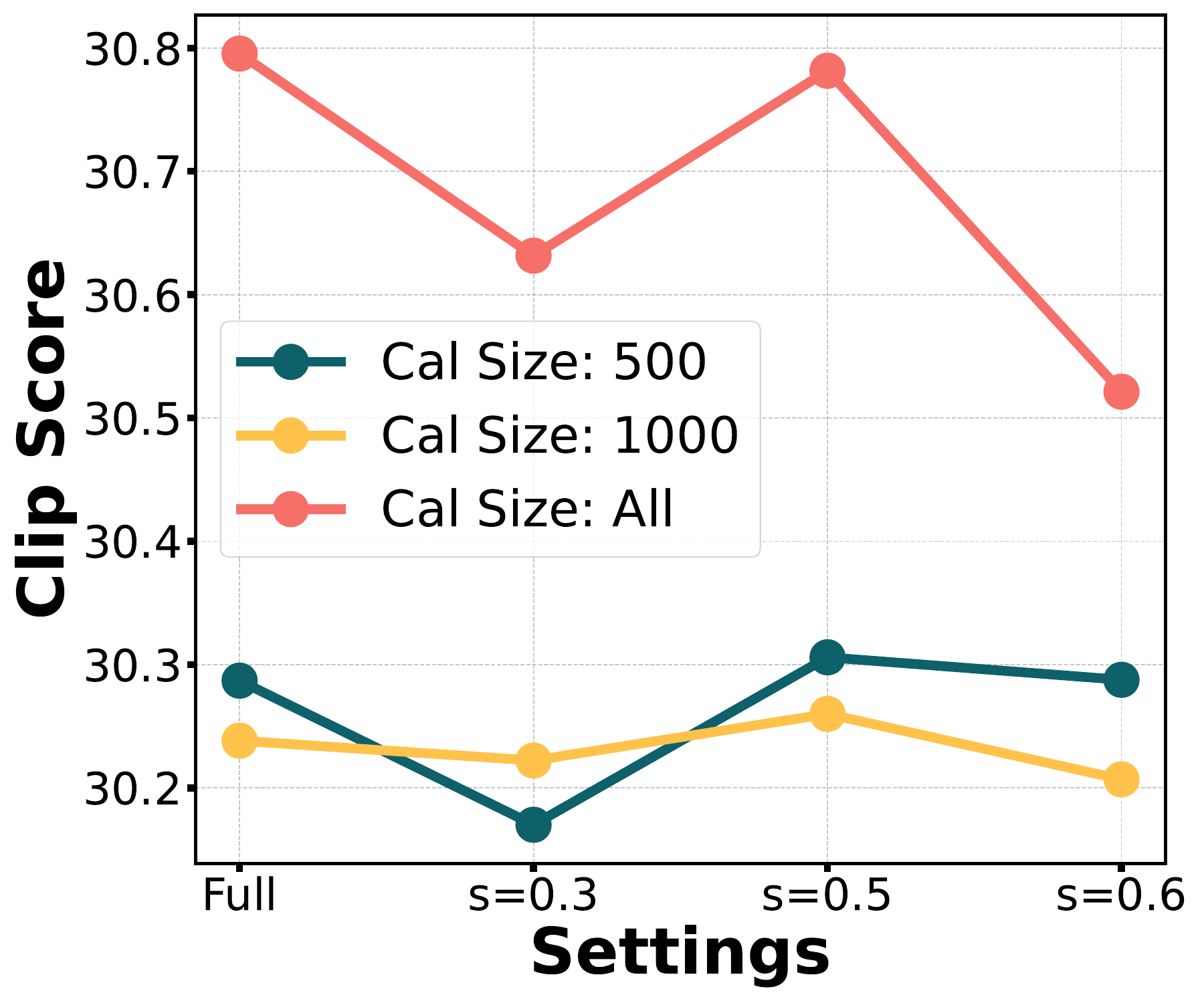}
    \caption{The achieved Clip Scores when using different calibration set sizes with varied $s$, where ``Full'' indicates no use of the mixed-resolution denoising.}
    \label{fig:cal_set}
    \vspace{-1em}
\end{figure}

\textbf{Observations.} As shown in \cref{fig:cal_set}, the performance trends for the small calibration sets align well with those for the full dataset. Specifically, a setting that achieves a higher Clip Score on a small calibration set is highly likely to achieve higher Clip Scores when using the full dataset. This strategy eliminates the cumbersome process of hyperparameter tuning on a large amount of data.

\begin{table}[t]  \centering
  \caption{Evaluation on Imagenet21K\_Recaption dataset. Latency is measured as the single-image generation time on an NVIDIA A5000 GPU for SD V1.5 and LCM, and an NVIDIA H100 GPU for PixArt-$\alpha$.}
  \addtolength{\tabcolsep}{3.0pt}
  \resizebox{0.49\textwidth}{!}{
      \begin{tabular}{c|cc|c}
          \toprule
          \textbf{Model}   & \textbf{FID} $\downarrow$ & \textbf{Clip Score} $\uparrow$ & \textbf{Latency} (s) $\downarrow$ \\
          \midrule
          SD V1.5  &   30.08 & 32.13 & 2.930 \\
           w/ PostDiff   &  27.57 & 31.75 & 1.139    \\  \midrule
          LCM   &  39.00 & 28.72 & 0.825 \\
             w/ PostDiff   &   37.74 & 28.22 & 0.651   \\  \midrule
            PixArt-$\alpha$  & 35.58 & 32.40 & 1.752  \\
           w/ PostDiff   & 32.69 & 32.04 & 1.382 \\  
          \bottomrule
      \end{tabular}  }   
      
      \label{tab:imagenet}
\end{table}

\section{Evaluation on the Imagenet21K\_Recaption Dataset}
\label{sec:imagenet}
In \cref{tab:imagenet}, we apply PostDiff to SD V1.5, LCM, and PixArt-$\alpha$ and evaluate the achieved performance on the Imagenet21K\_Recaption dataset\footnote{\url{https://huggingface.co/datasets/gmongaras/Imagenet21K_Recaption}}, where we randomly pick up 10,000 text prompts. 

The results show that applying PostDiff consistently improves the FID of different models while maintaining a comparable Clip Score and lower latency. This further demonstrates the robustness and generality of our proposed method across diverse text prompts and diffusion models.

\section{Combine PostDiff with Fine-Tuning}
\label{sec:finetune}
As a common practice, diffusion models are typically trained on a single resolution, which can  lead to relatively lower-quality outputs for low-resolution images. To address this, we experiment with fixed resolution schedules to fine-tune diffusion models, aiming to explore whether this approach can enhance the performance of our mixed-resolution strategy and, in turn, further improve PostDiff.

\textbf{Setup.}
We fine-tune SD V1.5 (the original version, as we were unable to obtain the dataset used for Dreamshaper's fine-tuning) and BK-SDM-Tiny \cite{kim2023architectural} (a light-weight version of SD V1.5) using the LAION-Art dataset\footnote{\url{https://huggingface.co/datasets/fantasyfish/laion-art}}. The batch size is set to 128, with 15,000 training iterations and a learning rate of 1e-5. We set $\beta=1/2$ and $s=1/2$ for SD V1.5, i.e., when the corresponding low-resolution steps (i.e., 501-1000) are randomly selected, half-resolution images are used to fine-tune the model, and $\beta=1/2$ and $s=1/5$ for BK-SDM-Tiny.

\textbf{Observations.}
As shown in \cref{tab:fine-tuning}, fine-tuning notably improves both FID and Clip Score, demonstrating the compatibility of our PostDiff with fine-tuning. Given that this process is fast and lightweight (about one day using one NVIDIA H200 GPU for SD V1.5), it provides a practical way to deploy PostDiff with enhanced performance while maintaining high generation efficiency.

\begin{table}[t]  \centering
  \caption{Performance of PostDiff w/o and w/ fine-tuning (FT).}
  \addtolength{\tabcolsep}{3.0pt}
  \resizebox{0.49\textwidth}{!}{
      \begin{tabular}{c|c|cc}
          \toprule
          \textbf{Model}  &\textbf{FT}  & \textbf{FID} $\downarrow$ & \textbf{Clip Score} $\uparrow$ \\
          \midrule
          \multirow{2}{*}{\textbf{SD V1.5}}          &       &  63.64   & 19.47 \\
                                                         &  \Checkmark     &  22.28   & 28.68   \\ \midrule
                                                
        \multirow{2}{*}{\textbf{BK-SDM-Tiny}}         &       &  19.87   & 25.87  \\
                                                        &  \Checkmark     &  19.38   & 28.44 \\ 
                                                 
          \bottomrule
      \end{tabular}  }   
      
      \label{tab:fine-tuning}
\end{table}

\section{Illustration of Hybrid Module Caching}
\label{sec:caching}

\begin{figure}
    \centering
    \includegraphics[width=0.95\linewidth]{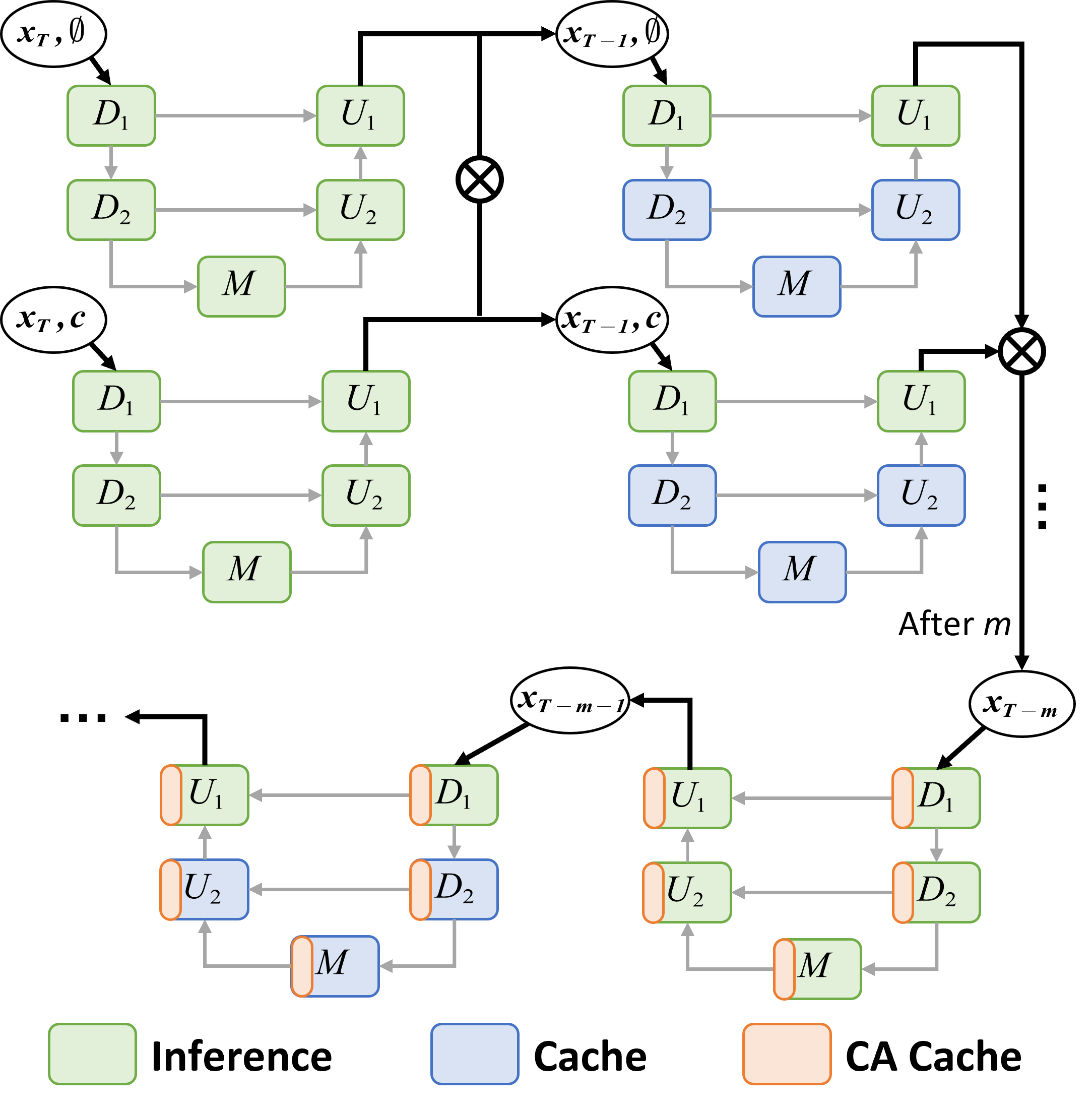}
    \caption{Overview of our hybrid module caching strategy.}
    \label{fig:cache_module}
    \vspace{-1em}
\end{figure}

In \cref{fig:cache_module}, we illustrate our hybrid module caching strategy, which effectively leverages feature redundancy across different timesteps, complementing the method description in Sec. 2.5 of our main paper.

\section{Comparisons with Prior Works} 
\label{sec:compare}
PostDiff adopts a post-training setting, i.e., it requires no fine-tuning and is easy to use. The effectiveness of mixed-resolution denoising in this setting stems from the distinct emphasis on low- and high-frequency components in the early and late denoising stages.
In contrast, Pyramid Flow \cite{jin2024pyramidal} interprets the diffusion trajectory as a multi-stage pyramid and trains all pyramid stages end-to-end from scratch or through model-specific fine-tuning, where the denoising process is explicitly adapted to the chosen resolution. The two works leverage different aspects of resolution.

DeepCache \cite{ma2024deepcache} is specifically tailored for U-Net, whereas PostDiff is applicable to both U-Net and DiT, achieving a 3.72 FID improvement with a $32.7\%$ latency reduction on PixArt-$\alpha$. Furthermore, neither DeepCache nor Faster Diffusion \cite{li2023faster} addresses redundant CFG during the late denoising phase. In contrast, PostDiff removes this redundancy to further boost efficiency, achieving $43.1\%$/$34.4\%$ latency reduction and improving FID by 1.2/1.1 compared to Faster Diffusion and DeepCache, respectively (see Tab. 3 of our main paper).

\section{More Visualization Results}
\label{sec:visual}

\begin{figure*}[!t]
    \centering
    \includegraphics[width=1\textwidth]{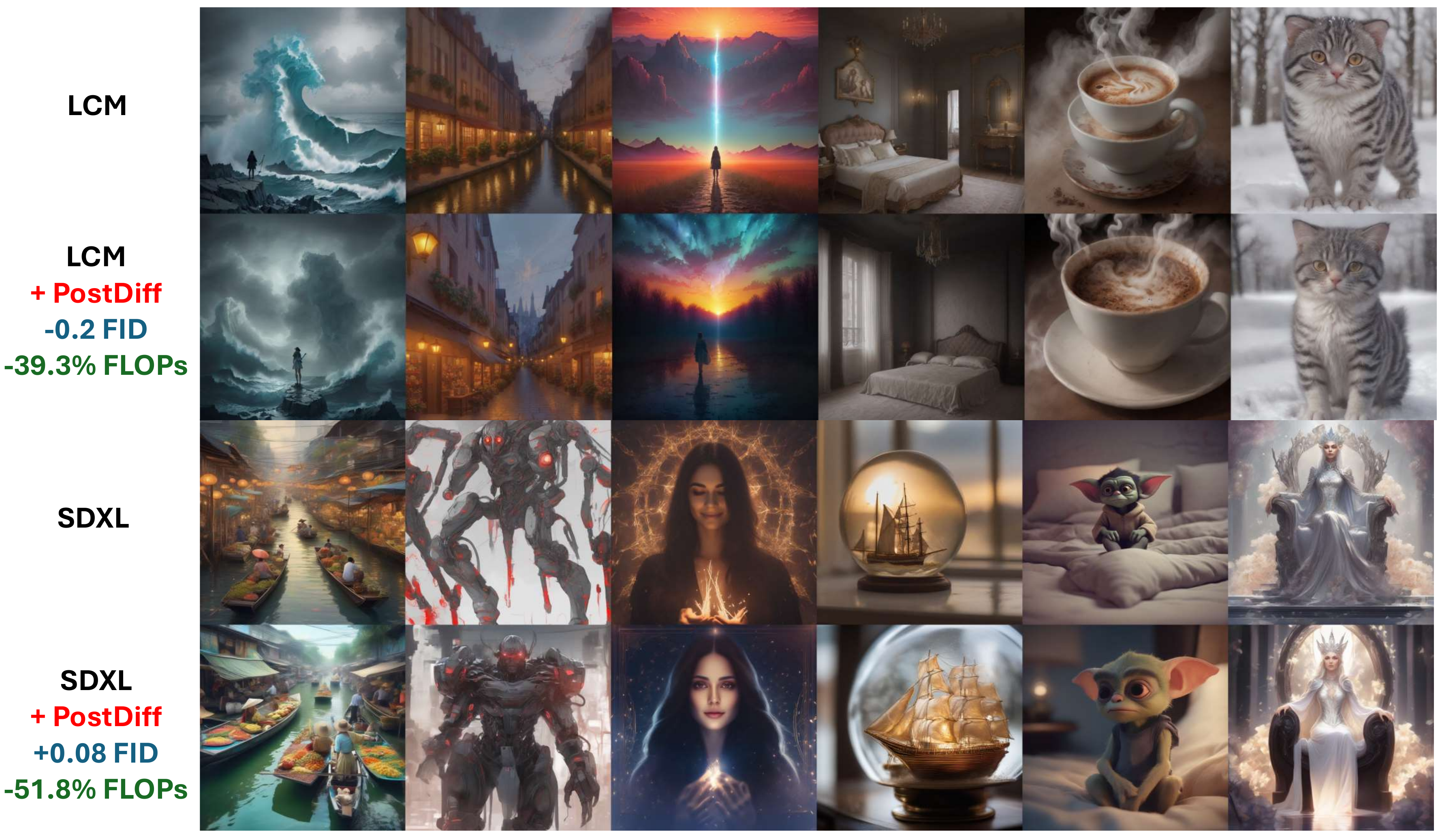}
    \caption{
        More visualization of generated images of different diffusion models w/o and w/ using our PostDiff.
        \label{fig:more_visualization}
        }
        \vspace{-3mm}
\end{figure*}

We visualize the text-to-image results generated by the original models alongside those produced using our PostDiff approach in \cref{fig:more_visualization}, as a complement to Fig. 6 of our main paper. As evidenced by the FID Scores, PostDiff achieves significant speedups while preserving details and, in some cases, further enhances visual quality.

\begin{figure*}
    \centering
    \includegraphics[width=0.85\linewidth]{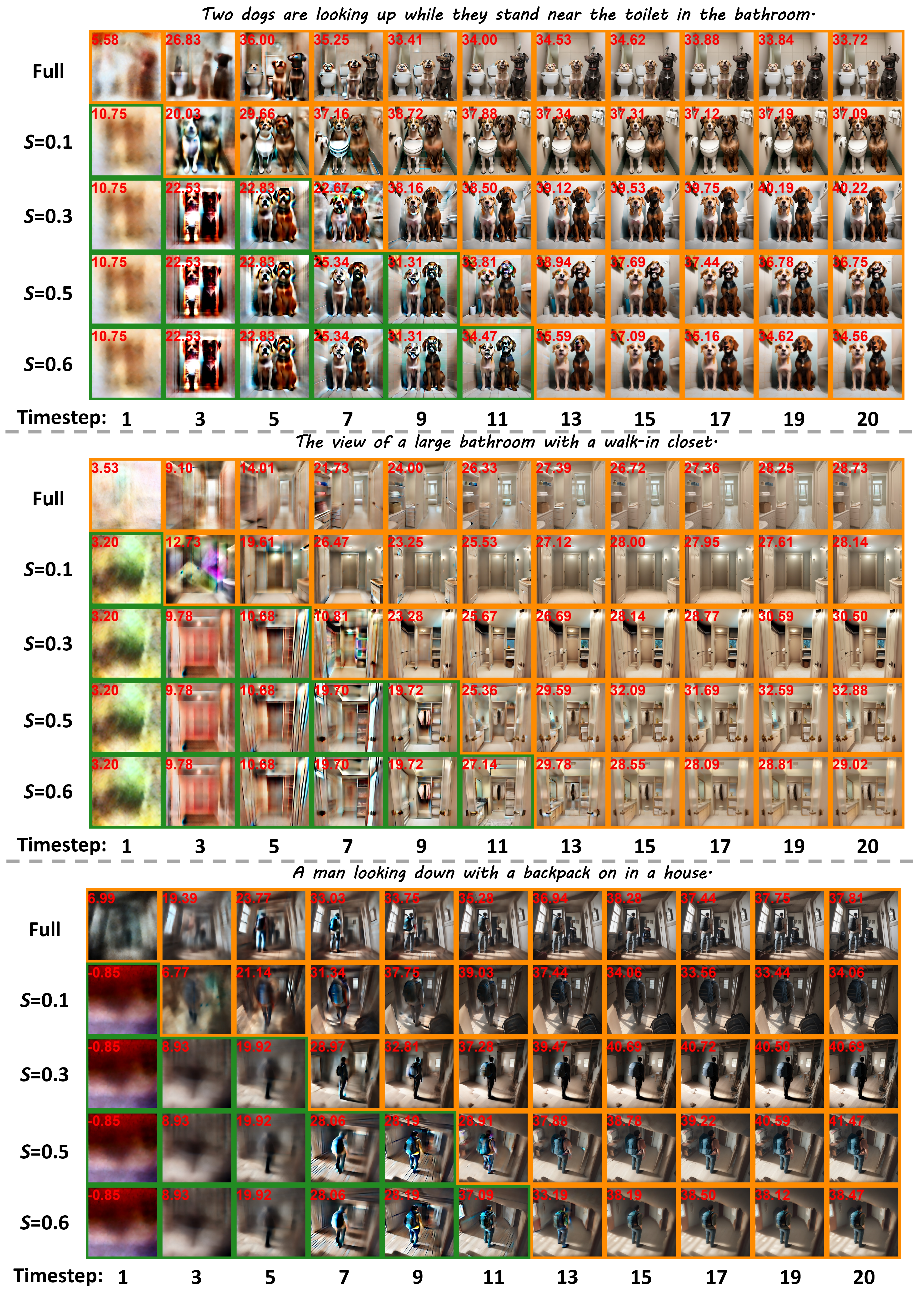}
    \caption{Visualize the Clip Score after each denoising step using full resolution (row 1) and mixed resolution (rows 2-5). The steps using low generation resolution are highlighted in green and those using high generation resolution are highlighted in orange.}
    \label{fig:mix}
    \vspace{-1em}
\end{figure*}

In addition, we include additional visual samples along with Clip Scores at each denoising step under different mixed-resolution settings in \cref{fig:mix}, as a complement to Fig. 3 of our main paper. It is consistently observed that utilizing low-resolution images to capture low-frequency components can potentially improve the final results.

\begin{figure*}
    \centering
    \includegraphics[width=0.95\linewidth]{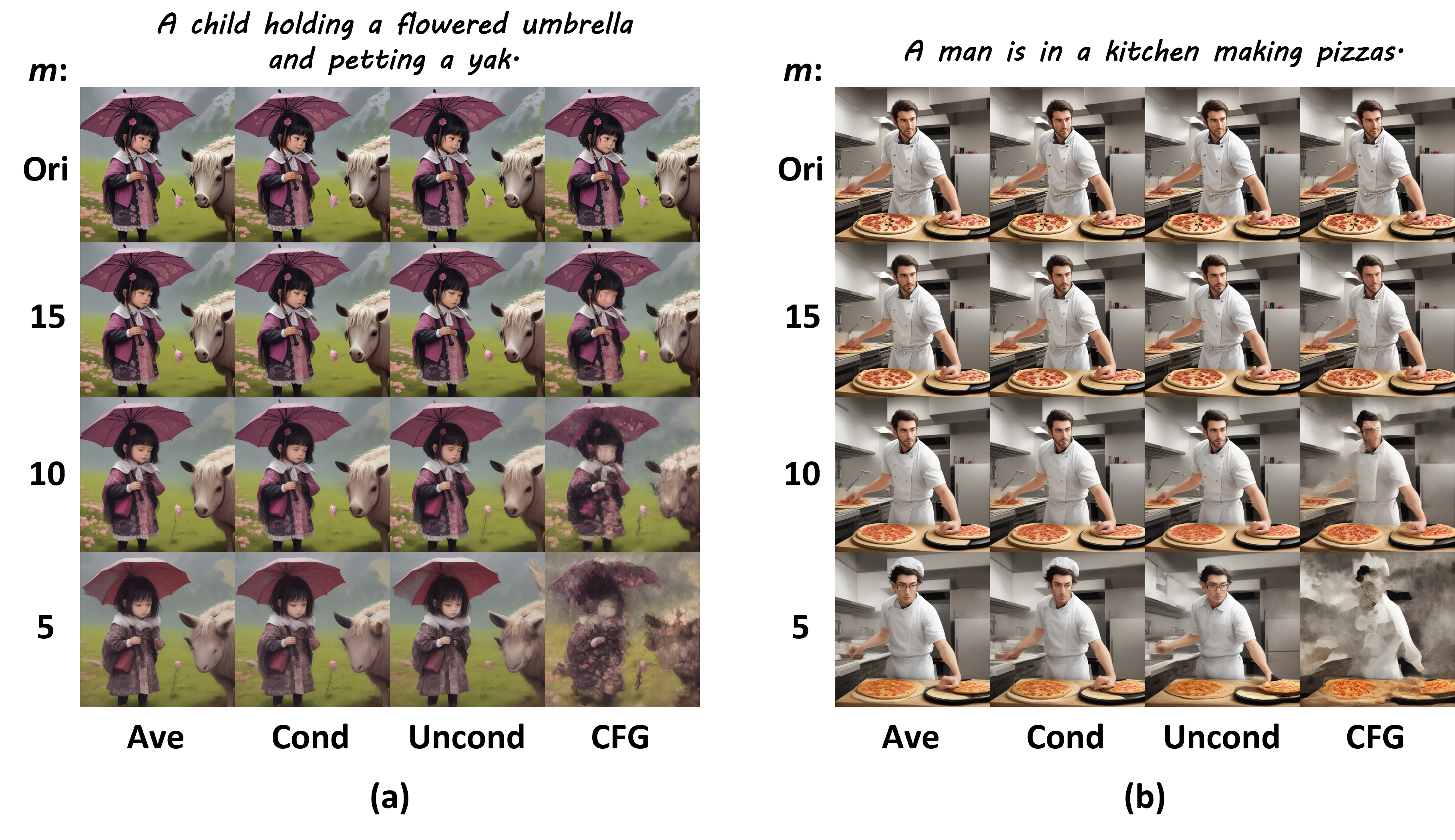}
    \caption{Generation results using different choices of $CA_{\text{cache}}$ with varying $m$. ``Ori'' indicates no use of cross-attention cache.}
    \label{fig:cache_choice}
    \vspace{-1em}
\end{figure*}

Furthermore, we also provide additional visual examples showcasing the impact of different choices for \( CA_{\text{cache}} \) with varying \( m \) values in \cref{fig:cache_choice}, as a complement to Sec. 2.5 of our main paper. The ``Cond'' choice generally performs best, maintaining better consistency with the prompt.

\section{Prompts for Text-to-Image Generation}
\label{sec:prompt}

In this section, we provide the prompts we used to generate images in the main paper and supplementary material.

\subsection{Figure 6 in the Main Paper}
\subsubsection{SD V1.5}
\begin{enumerate}
    \item A mystical underwater scene with glowing coral
    \item A bustling cyberpunk metropolis at night, illuminated by a kaleidoscope of neon lights and holographic advertisements. The streets are crowded with people wearing futuristic attire.
    \item A forest with glowing mushrooms and creatures.
    \item A fantasy castle on a hill surrounded by clouds.
    \item A snowy mountain landscape with a cozy cabin and smoke coming from the chimney.
    \item Imagine and detail very clearly: the exciting rebirth of an energetic being in the vast void of space, together with a glorious phoenix. It begins with a stellar background that dazzles with the beauty of the cosmos, with brilliant nebulas and resplendent constellations.
\end{enumerate}

\subsubsection{PixArt-$\alpha$}
\begin{enumerate}
    \item A stunning Japanese-inspired fantasy painting of a lone samurai, silhouetted against a massive full moon, standing beneath a windswept, crimson-leafed tree. Falling petals swirl around him, creating a melancholic yet serene atmosphere. The dramatic chiaroscuro lighting highlights the dramatic contrast between the cool-toned background of deep blues and grays and the warm reds of the foliage. This captivating scene is reminiscent of Yoshitaka Amano's work, with the dramatic lighting of Ivan Shishkin.
    \item MysticSplash. Ink-splash-style. Ink-splash-style. Extreme closeup of a dapper figure in a stylized, richly detailed black top hat, adorned with decorative golden accents, stands against a white background. The character is a skeleton with very detailed skull and long canines as vampire fangs. He cloaked in a vibrant victorian jacket, featuring intricate golden embellishments and a deep red vest underneath. He wears a large victorian monocle with a yellow-tinted lense and copper frame very reddish. Exquisite details include a shiny silver cross and a blue gem on the chest, harmonizing with splashes of paint in vivid hues of blue, gold, and red that artistically cascade around the figure, blending an impressionistic flair with elements of surrealism. The atmosphere is whimsical and opulent, evoking a sense of grandeur and mystery.
    \item An ancient, overgrown temple in a dense jungle, illuminated by the soft light of early morning.
    \item The image is a landscape photograph of a mountain range with a river flowing through it. The river is surrounded by a rocky shoreline with small pebbles and boulders. The water is a deep blue color and reflects the mountains and trees in the water. The mountains in the background are tall and imposing, with a mountain peak in the distance. The sky is clear and blue, and the sun is shining brightly, creating a beautiful reflection of the mountains on the water's surface. The overall mood of the image is peaceful and serene. mad-sprklngtr, flrlizer.
    \item A portrait of a cybernetic geisha, her face a mesmerizing blend of porcelain skin and iridescent circuitry. Her elaborate headdress is adorned with bioluminescent flowers and delicate, glowing wires. Her kimono, a masterpiece of futuristic design, shimmers with holographic patterns that shift and change, revealing glimpses of the complex machinery beneath. Her eyes gaze directly at the viewer with an enigmatic expression.
    \item A single, crazy blue and black spaceship in the sky. It overwhelms the viewer with its artistic flying skills while trailing a meteor tail. Ace pilot of the Republic who was unrivalled in the 1940s. His second name is: The Magician of the Blue Wings, a genius aviator, one of a kind in 100 years. The warriors who challenged him, were destroyed by him, were overrun by him and scattered became many stars. The Milky Way is said to be the graveyard of such aerialists.
\end{enumerate}

\subsection{Figure \ref{fig:more_visualization} in the Supplementary Material}
\subsubsection{LCM}
\begin{enumerate}
    \item A massive wave crashing onto a rocky shore, with a lone figure standing defiantly against the storm, holding a glowing staff.
    \item A whimsical town where all the buildings are made of candy, with rivers of chocolate and lollipop streetlights glowing faintly in the dusk.
    \item The long journey home, vibrant glow.
    \item Parisian luxurious interior penthouse bedroom, dark walls, wooden panels.
    \item Novuschroma style cup of coffee with swirling steam.
    \item Scottish fold kitten, professional photo, in snow, high detail, close-up view, quantum rendering, masterpiece, professional photo.
\end{enumerate}

\subsubsection{SDXL}
\begin{enumerate}
    \item Floating market of old Bangkok by day, atmospheric lighting, awesome background, highly detailed, cinematicfantasy, dreaming, best quality, double exposure, realistic, whimsical, fantastic, splash art, intricate detailed, hyperdetailed, maximalist style
    \item A cybernetic warrior with mechanical arms and glowing red eyes.
    \item A smiling beautiful sorceress with long dark hair and closed eyes wearing a dark top surrounded by glowing fire sparks at night, symmetrical body, symmetrical face, symmetrical eyes, magical light fog, deep focus+closeup, hyper-realistic, volumetric lighting, dramatic lighting, beautiful composition, intricate details, instagram, trending, photograph, film grain and noise, 8K, cinematic, post-production.
    \item Miniature sailing ship sailing in a heavy storm inside of a horizontal glass globe inside on a window ledge golden hour, home photography, 50mm, Sony Alpha a7.
    \item Little cute gremlin sitting on a bed at night thinking about the world, cinematic, muted colors, faded, by pixar and dreamworks.
    \item A regal elf queen sitting on a crystalline throne, her gown shimmering like liquid silver, with a crown of glowing flowers.
\end{enumerate}


\end{document}